\documentclass[sigconf]{acmart}
\usepackage{subcaption}
\usepackage{tabularx}
\usepackage{multirow}
\usepackage{xspace}
\usepackage{lscape}
\usepackage{amsmath}
\usepackage{capt-of}% or \usepackage{caption}
\usepackage{booktabs}
\usepackage{varwidth}
\usepackage{pifont}
\usepackage{amsmath,amsfonts}
\usepackage{algorithmic}
\usepackage{graphicx}
\usepackage{url}
\usepackage{enumitem}

\AtBeginDocument{%
  }

\setcopyright{acmlicensed}
\copyrightyear{2026}
\acmYear{2026}
\acmDOI{XXXXXXX.XXXXXXX}
\acmConference[SIGSPATIAL'26]{The 34th ACM SIGSPATIAL International Conference on Advances in Geographic Information Systems}{November 03--06, 2026}{Riverside, CA, USA}
\acmISBN{978-1-4503-XXXX-X/26/11}

\newcommand{\ArtifactURL}{To be released}
\newcommand{\MapArtifactURL}{To be released}
\newcommand{\MetricsArtifactURL}{To be released}
\newcommand{\TrafficSimArtifactURL}{To be released}
\newcommand{\parisData}{GreaterParis dataset}

\newcommand{\parisDataShort}{GreaterParis}

\definecolor{lightergray}{gray}{0.90}

\newcommand\greybox[1]{%
  \par\noindent\colorbox{lightergray}{ 
    \begin{minipage}{0.465\textwidth}#1\end{minipage}%
  }%
  \vskip\baselineskip%
}

\begin{document}

% \title{Validating the Physics of Synthetic Movement: A Framework for Evaluating Macroscopic Human Mobility in LLM-Driven Simulators} % Só Exemplo.
%A Framework for Validating LLM-Based Human Mobility \\or\\
\title{When Plausible Is Not Realistic: Evaluating Human Mobility in LLM-Based Urban Simulation}
% Do social LLM-based simulators represent mobility realistically?

%\author{Gustavo H. Santos$^{1,2}$, Aline Carneiro Viana$^1$, Thiago H. Silva$^{2,3}$}
%\affiliation{$^1$\institution{Inria}\country{France},\\ $^2$\institution{Inria}\country{France}}
%  \vspace{0.2em}
  %$^1$Inria, France \hspace{0.5cm} $^2$Universidade Tecnologica Federal do Parana, Brazil \hspace{0.5cm}  $^3$ University of Toronto, Canada } 

 \author{Gustavo H. Santos}
 \authornote{Corresponding author: gustavohenriquesantos@alunos.utfpr.edu.br}
 \affiliation{%
   \institution{UTFPR, Brazil and} \country{Inria, France}
   %\city{Curitiba}
  %\country{Brazil}
 }

 \author{Aline Carneiro Viana}
\affiliation{%
   \institution{Inria, France} \country{}
 }

 \author{Thiago H. Silva}
 \affiliation{%
   \institution{UTFPR, Brazil and} \country{U. of Toronto, Canada}
 }

\renewcommand{\shortauthors}{Santos et al.}

\begin{abstract}
LLM-based generative agents are increasingly used in urban simulators, yet it remains unclear whether they reproduce empirically realistic human mobility patterns or merely generate plausible mobility narratives. We introduce a validation framework for evaluating the mobility of generative agents of LLM-based urban simulators against real-world mobility data. For this, we use mobility laws, temporal rhythms, network motifs, semantic activity transitions, and behavioral mobility profiles. Using datasets from the Greater Paris region and Shanghai, we evaluate \textit{AgentSociety} and \textit{CitySim} across multiple dimensions of mobility realism. Our analysis reveals a substantial gap between narrative plausibility and empirical mobility realism. Although the simulators capture some high-level semantic activity distributions, they struggle to reproduce core spatial and temporal constraints, including realistic trip-length distributions, origin-destination flows, dwell times, and transition dynamics. We further observe that realistic mobility diversity is unstable across default prompting configurations and may require explicit profile-aware initialization. To support reproducible evaluation, we also contribute scalable and open LLM-driven infrastructure for regional-scale map generation, observability-enhanced simulation, mobility-metric computation, and traffic simulation. Our findings highlight the need for rigorous empirical validation of LLM-based urban simulators and provide practical tools for building more realistic and reproducible urban simulation systems.
\end{abstract}




\received{30 May 2026}
\received[revised]{x March 2026}
\received[accepted]{x June 2026}

\maketitle

\section{Introduction}

Large Language Models (LLMs) are increasingly used to simulate urban behavior and human mobility \cite{liang2024exploring,zeng2025crimemindsimulatingurbancrime,sung2026planning}. Recent frameworks such as \textit{AgentSociety} \cite{agentsociety} and \textit{CitySim} \cite{citysim25} support generative agents operating in simulated urban environments, enabling them to plan routines, engage in social interactions, and adapt to contextual information, as weather, work schedules, and personal needs.
These systems suggest a new paradigm for urban simulation, but raise a central question: \textit{Do synthetic agents reproduce empirical human mobility laws, or only generate trajectories that appear plausible?}

This distinction is fundamental for simulation validity. Decades of mobility research characterized the complexity of human behavior, the structured circadian routines, and the novelty-seeking behavior of individuals, identifying patterns such as truncated power laws in \textit{travel distance} \cite{Gonzalez2008}, predictable visitation dynamics \cite{Song2010Entropy,teixeira:hal-03360537}, recurrent topological \textit{mobility motifs} \cite{Schneider2013}, and heterogeneous yet not-random \textit{mobility profiles} \cite{licia_sig2020,and18jan,Pappalardo:2015}.
Traditional mobility simulators are routinely validated against these empirical patterns before being applied to policy analysis or forecasting \cite{computable_city24,Sargent_val_sim_13,Epstein2007-mx,validation_simulation95}. 
In contrast, LLM-driven simulators are typically evaluated through plausibility-oriented assessments, focusing on whether generated schedules or behaviors appear coherent and believable \cite{agentsociety,citysim25}. 

While such evaluations may capture temporal or narrative consistency, they do not establish realistic mobility behavior. Indeed, plausibility-based mobility evaluations primarily capture a form of ``face validity''~-- i.e., whether generated behavior appears reasonable or believable~-- rather than ``objective mobility realism'': an agent's mobility may appear plausible (waking up at 7:00, going to work, having lunch at noon, returning to home at 18:00 every week day), while the underlying mobility process may still fail to reproduce empirical human mobility dynamics (e.g., exhibiting unrealistic travel distances, random visitation frequencies, incorrect exploration dynamics, or the absence of spatial confinement effects). \textit{We define mobility realism as the joint reproduction of spatial visitation dynamics and temporal activity organization according to empirical human mobility regularities.} 
However, evaluations of existing LLM-driven simulators generally lack systematic validation of these coupled spatio-temporal properties \cite{larooij2025largelanguagemodelssolve, Kapp2023, gao2026llmshumanmobilityopportunities}. 

We introduce \textit{a comprehensive validation framework for LLM-based urban simulators grounded in real-world mobility data and seminal literature on human mobility laws.} To the best of our knowledge, this is \textit{the first systematic evaluation of LLM-based urban simulators against established empirical mobility laws and spatio-temporal behavioral metrics} of human mobility. Our findings highlight the need for rigorous empirical validation as a standard practice alongside plausibility-oriented assessments. %Besides, 
\textit{We release a set of artifacts providing a foundation for building more realistic and reproducible urban simulation systems}. Overall, we make three contributions:

\noindent\textbf{Methodological:} After a detailed literature review (cf. \textbf{\S \ref{secRelatedWork}}), we propose a modular framework for LLM-driven mobility realism evaluation (cf. \textbf{\S \ref{sec:methodology}}) across five dimensions: spatial mobility laws, temporal rhythms, topological motifs, behavioral profiles, and semantic/spatio-temporal activity patterns. Together, these components assess whether simulated trajectories reproduce empirical human mobility regularities across spatial and temporal dimensions. \vspace{0.1cm}
 
\noindent\textbf{Empirical:} Using pseudoanonymized mobility datasets from Greater Paris and Shanghai, we evaluate \textit{AgentSociety} \cite{agentsociety} and \textit{CitySim} \cite{citysim25} beyond narrative plausibility. For this, we introduce \textit{En-AgentSociety}, an enhanced version of \textit{AgentSociety} with improved mobility traceability, observability, and scalability.
Across metrics, we find a consistent gap between plausible agent behavior and empirically valid mobility (\textbf{\S \ref{sec:results}}): Simulators reproduce some high-level semantic/activity distributions but fail to recover key spatial constraints, displacement patterns, visitation structures, and transition dynamics. We show that behavioral diversity cannot be achieved through generic persona prompting alone, but requires profile-aware initialization and stronger spatial constraints.

\noindent\textbf{Artifact:} Methodological and empirical contributions compose a released infrastructure for reproducible large-scale evaluation of LLM-driven mobility simulations\footnote{\ArtifactURL. Anonymous for the revision phase}. The stack includes regional-scale map generation, an observability-enhanced fork of \textit{AgentSociety}, scalable Rust+Python metric computation,  and an open reimplementation of the traffic simulator \textit{AgentSociety} and \textit{CitySim} depend on. Together, they support standardized benchmarking of LLM-based mobility simulators at regional scale.

We discuss paths toward higher-fidelity LLM-based urban simulation in \textbf{\S \ref{sec:discussion}} and conclude in \textbf{\S \ref{sec:conclusion}}.

\section{The Human Mobility Landscape}
\label{secRelatedWork}

 Realistic city-scale mobility simulation supports data-driven urban services, but its reliability depends on accurately reproducing human mobility patterns. This section reviews empirical mobility regularities and limitations of recent LLM-based urban simulators.

\subsection{Empirical Human Mobility Patterns}

Human mobility exhibits strong statistical regularities across spatial and temporal scales. Using mobility traces from more than 100,000 mobile-phone users, González et al.~\cite{Gonzalez2008} showed that human \textit{travel distance} follows truncated power-law distributions and that individual movements remain spatially bounded according to a characteristic \textit{radius of gyration}. Subsequent work by Song et al.~\cite{Song2010Entropy} demonstrated that human trajectories are highly predictable, estimating an upper-bound \textit{predictability} of approximately 93\% despite population-scale behavioral diversity.

Prior studies also revealed strong regularities in daily mobility structure. Schneider et al.~\cite{Schneider2013} showed that a small set of recurrent \textit{mobility motifs} can represent up to 90\% of weekday mobility patterns, while the number of \textit{daily visits} approximately follows a log-normal distribution. More recently, Schläpfer et al.~\cite{Schlapfer2021-pw} explored the joint relationship between \textit{travel distance} and \textit{visitation frequency}, showing that frequent short-range trips and infrequent long-range trips emerge at different spatial scales~\cite{Alessandretti2020-uo}, from neighborhoods to metropolitan regions, with important implications for processes such as disease spreading~\cite{Heine2023-sd}.
\greybox{\textbf{Need:} \textit{  Human mobility is governed by robust empirical laws rather than arbitrary movement behavior. Consequently, urban simulators based on LLM-driven agents should be evaluated not only by the plausibility of generated narratives, but also by their ability to reproduce these established mobility patterns.} \vspace{-0.1cm}
}

\subsection{Behavioral and Semantic Mobility}

Besides exhibiting strong regularities shaped by routine behaviors, human mobility also includes periods of exploration that vary substantially across individuals. Prior work proposed metrics such as \textit{entropy}, \textit{regularity}, \textit{stationarity}, and \textit{diversity} to characterize these differences~\cite{Teixeira2019,Teixeira2021}. Building on these measures, Amichi et al.~\cite{licia_sig2020} identified three major \textit{mobility profiles}: \textit{Scouters}, who frequently explore new locations; \textit{Routiners}, who primarily revisit familiar places; and \textit{Regulars}, who exhibit intermediate behavior.  These behavioral distinctions have important implications for mobility prediction and simulation~\cite{licia2021,amichi:hal-03905517,esper_sig24,kouam2023zen,kouam:hal-05248595}.

\greybox{\textbf{Need:}\textit{ Realistic simulators should capture both routine-like mobility laws and heterogeneous exploration patterns across individuals.
}} \vspace{-0.3cm}
Beyond behavioral diversity, realistic urban simulation must also capture the semantic context underlying mobility decisions. While Location-Based Social Network (LBSN) datasets present known demographic and platform biases~\cite{checkin_badges_2021}, they provide valuable semantic information such as POI categories and activity transitions that are often absent from traditional mobility datasets. Recent mobility datasets and LLM-based urban simulators increasingly incorporate this semantic dimension through activity types, transportation modes, and spatio-temporal activity analysis~\cite{chasse:hal-05365088,wang2024largelanguagemodelsurban}. 
\greybox{\textbf{Need:}
\textit{The semantically rich trajectories generated by LLM-based simulators require evaluation metrics that capture semantic and spatio-temporal dynamics beyond spatial statistics.}}

\subsection{LLM-Based Urban Simulators}

Recent LLM-based urban simulators aim to reproduce human behavior through large-scale generative agents:

\noindent\textbf{\textit{AgentSociety}~\cite{agentsociety}:} an open-source, OpenStreetMap-based framework combining LLM-driven decision-making and psychologically inspired behavioral modules~-- including economic, cognitive, planning, mobility, and social components~-- for urban trajectory generation. After a detailed framework review and the enhancements we brought to the \textit{AgentSociety} simulator (cf. \S \ref{sec:methodology_simulator}), we designed Fig.~\ref{fig:simulation_diagram} to illustrate the simulation pipeline and module interactions. At initialization (cf. Fig. \ref{fig:simulation_diagram}), the \textit{EconomyBlock} and \textit{NeedsBlock} assign demographic, financial, and behavioral attributes to each agent, including occupation, income, consumption level, and dynamic needs such as hunger, energy, safety, and social connection. Needs evolve over time according to Maslow’s Hierarchy~\cite{Maslow1943} and influence subsequent planning decisions. Daily behavior is generated through the \textit{PlanBlock}, which uses LLM calls guided by the Theory of Planned Behavior~\cite{theory_planned_behavior}. Based on the agent’s current state (e.g., location, time, weather, emotions, occupation, and dominant need), the simulator first selects a high-level behavioral strategy and then generates a sequence of fine-grained actions associated with specific execution modules. The \textit{DispatcherBlock} routes each action to the corresponding simulator component.

The \textit{MobilityBlock} governs movement decisions and destination selection. For each movement step, the LLM selects a POI category, subtype, and exploration radius conditioned on the agent’s current plan and contextual state. Candidate POIs are then sampled using a Density-aware Gravity model based on Distance-decayed Attraction: $w_i = \frac{\rho_k}{d_i^2}$, where $\rho_k$ represents the POI density within spatial ring $k$, and $d_i$ is the distance to candidate POI $i$. Selection probabilities are computed by normalizing these weights over the candidate set. \textit{This mobility-selection mechanism is central to our later analysis because many observed spatial mismatches originate from failures in POI selection and movement generation.}

At the end of each simulation step, the \textit{CognitionBlock} updates the agent’s internal state, including emotions, memories, and contextual reasoning, while the \textit{OtherBlock} estimates durations for actions not directly handled by specialized modules. The \textit{SocialBlock} manages interactions between agents using a weighted social graph. Because \textit{AgentSociety does not generate this graph automatically and modeling realistic social ties introduces additional complexity, we disable this component in our experiments and initialize agents with empty social networks.}
Finally, agents maintain a temporal memory structure recording observations, locations, and prior actions, allowing previous experiences to influence future decisions/plans.

\noindent\textbf{\textit{CitySim}~\cite{citysim25}:} an extended, closed-source version of \textit{AgentSociety} with richer cognitive, spatial, and social mechanisms aimed at improving behavioral realism. In addition to demographic attributes, agents are initialized with personality traits, preferences, and long-term goals that evolve according to need fulfillment and life context. The framework also introduces a spatial-memory system that stores subjective beliefs about previously visited POIs, influencing future destination choices. Compared with \textit{AgentSociety}, \textit{CitySim} employs more detailed planning and mobility-selection strategies. Daily schedules are generated using fine-grained time blocking, while destination selection combines distance-based attraction with memory-aware preferences and contextual reasoning. The framework also incorporates transport-mode selection and dynamically evolving social interactions through weighted social networks.

\greybox{\textbf{Need:}
\textit{\textit{AgentSociety} and \textit{CitySim} exhibit promising qualitative and semantic behaviors, but their evaluation remains limited with respect to established empirical mobility laws and behavioral mobility patterns. Existing assessments primarily focus on the plausibility and coherence of generated activities rather than on whether simulated trajectories reproduce real-world spatial, temporal, and topological mobility dynamics \cite{citysim25,agentsociety}. This gap motivates the need for systematic mobility-grounded validation of LLM-based urban simulators.}}

\section{Mobility Realism Evaluation Framework}
\label{sec:methodology}

Evaluating the realism of human mobility patterns in urban simulators requires (i) 
a broad set of metrics to quantitatively characterize urban dynamics and (ii) complete traceability and interpretability of generated mobility behaviors. However, current simulators lack the capabilities necessary for rigorous mobility evaluation. 
This section first presents the design enhancements introduced into the studied simulators, followed by the analytical framework and metrics used to evaluate mobility realism.

\subsection{Mobility Traceability and Observability Enhancement}
\label{sec:methodology_simulator}

\begin{figure}[htb]
    \centering
    \includegraphics[width=1\linewidth]{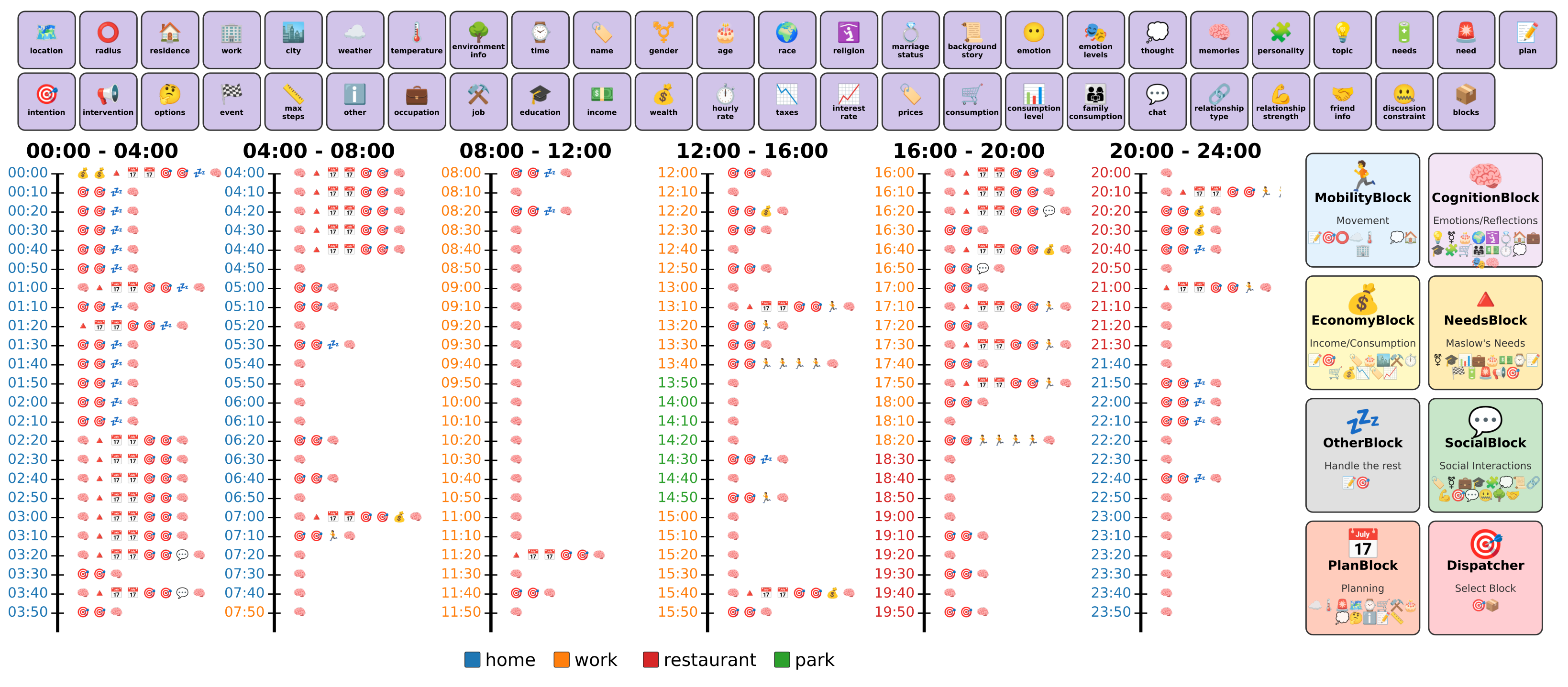}\vspace{-0.3cm}
   \caption{Overview of the \textit{AgentSociety} simulation pipeline and LLM calls modules during a daily mobility routine. Timestamped rows correspond to a 10-minute simulation step: e.g, at 07:10, the diagram shows two \textit{DispatcherBlock} calls, one \textit{MobilityBlock} call, and one \textit{CognitionBlock} call. These correspond to dispatching the agent's plan to the appropriate module (i.e., commuting to work), executing the mobility action, and updating the agent's internal state.
   }
    \label{fig:simulation_diagram}
\end{figure}

\textit{AgentSociety}~\cite{agentsociety} lacks several features essential for rigorous validation, particularly in the context of mobility modeling:
\begin{itemize}[leftmargin=*]
    \item[-] \textbf{Lack of Traceability:} \textit{AgentSociety} does not record visited POIs or their categories, which obscures the underlying location-selection mechanism and prevents verification of whether destination choices are genuinely needs-driven or effectively random.
    \item[-] \textbf{Limited Transparency and Observability:} \textit{AgentSociety} does not log block execution order or prompt-response pairs (lack of traceability), limiting reproducibility and systematic analysis of agent decision-making.
    
    \item[-] \textbf{Unconstrained POI selection:} Although \textit{AgentSociety} uses LLM-driven planning to determine daily activities and select POI categories, we observed that needs-driven mobility actions could select destinations from the full set of POIs available in the simulated map, rather than from POIs semantically consistent with the active need and spatially plausible given the agent's current context. As a result, agents with low hunger satisfaction could be routed to semantically unrelated or distant POIs, producing unrealistic trip distances, dwell patterns, and activity transitions. This behavior motivated our enhancements for recording selected POIs, POI categories, and mobility-decision traces.
\end{itemize}
\noindent \textbf{\textit{AgentSociety} enhancement.} To address these limitations, we developed \textit{En-AgentSociety}, an enhanced version of \textit{AgentSociety} that improves mobility traceability, simulator observability, and scalability. Fig.~\ref{fig:simulation_diagram} presents the main modules and the information flows of \textit{En-AgentSociety} for a simulated daily mobility routine. Several lower-level components are consolidated into eight high-level modules shown on the right side of the figure. The central timestamped workflow shows, at each 10-minute simulation step, the sequence of modules involved in an agent's decision-making, while the sequence of module calls illustrates the information exchanged during the simulation process. The figure also summarizes the 49 agent attributes used throughout the simulation, displayed at the top. The workflow is annotated with timestamps that are color-coded according to the agent's location (e.g., home or work). It is worth highlighting that the visualization of the timestamped workflow, agent locations over time, and module interactions is enabled by the traceability and observability enhancements described below.

\begin{itemize}[leftmargin=*]
    \item[-] \textbf{Traceability:} refers to information collection and storage. The enhanced framework logs complete mobility traces (visited locations and POIs), execution traces, and prompt–response pairs, enabling systematic analysis of agent decisions and simulator behavior. Logs are stored in a Dockerized ClickHouse database, with an optional local DuckDB backend for lightweight execution. Since prompt–response logs can be very large, detailed logging can be disabled through the simulator configuration.
    \item[-] \textbf{Observability:} refers to monitoring, statistics, latency analysis, and performance diagnosis. Building on the collected logs, the enhanced framework integrates an LGTM observability stack to monitor resource utilization, identify prompt-level latency bottlenecks, and provide aggregated execution statistics to support optimization efforts (cf. \textbf{\S\ref{sec:discussion}}).
    \item[-] \textbf{Scalability:} \textit{AgentSociety} depends on MOSSTool~\cite{MOSSTool} to generate simulation maps from OpenStreetMap data. While suitable for city-scale environments, the original implementation does not scale efficiently to large regions due to costly spatial matching and redundant serialization. To address these limitations, we implemented spatial indexing and batch-based processing, enabling scalable regional map generation. We release this optimized implementation as a public artifact \footnote{\MapArtifactURL. Anonymous for the revision phase}.
    Additional scalability benchmarks are reported in the Appendix \ref{sec:appdx_mosstool_benchmark}.
\end{itemize}

\noindent\textbf{\textit{CitySim} implementation.} 
Because \textit{CitySim}~\cite{citysim25} is a non-public extension of \textit{AgentSociety} (v1.5), we manually implemented its documented functionalities based on the descriptions provided in the associated publications. The resulting reconstruction was integrated into the same \textit{En-AgentSociety} artifact, enabling a consistent comparison between \textit{AgentSociety} and \textit{CitySim} under a common simulation infrastructure.
We do not provide a dedicated \textit{CitySim} illustration because its execution workflow follows the same overall structure as that shown in Fig.~\ref{fig:simulation_diagram}, albeit with additional agent attributes and decision-making blocks.
Notably, the unconstrained POI selection mechanism inherited from \textit{AgentSociety}
was corrected, reducing trip-length discrepancies and improving behavioral consistency. Nevertheless, the resulting mobility patterns remain substantially different from the real-world mobility baseline, likely due to the increased complexity of the simulator and its decision-making process (cf. \textbf{\S \ref{sec:results}}). Additional implementation details for \textit{CitySim} are provided in Appendix~\ref{sec:appdx_citysim}.

\subsection{Analytical Framework}
\label{sec:methodology_metrics}

For metrics with well-established parameterized formulations, we retain their original definitions as proposed in the literature. The complete mathematical expressions and computation procedures are reported in Table~\ref{tab:methodology_metrics} in Appendix \ref{sec_appendix_metrics}. For metrics whose parameter choices require further justification or adaptation to the \textit{AgentSociety} context, we provide a detailed discussion of the chosen parameter values and their rationale in the following.

\begin{itemize}[leftmargin=*]

\item \textit{Travel distance ($\Delta r$)}~\cite{Gonzalez2008} follows the truncated power law $P(\Delta r) = (\Delta r + \Delta r_0)^{-\beta} \exp(-\Delta r/\kappa),$ where $\Delta r$ is the distance between consecutive locations, $\beta$ is the scaling exponent, $\Delta r_0$ regularizes short distances, and $\kappa$ is the exponential cutoff representing the characteristic travel limit of the population. In \textbf{\S \ref{sec:results}}, we use the classical reference values for travel distance: $\beta = 1.75$, $\Delta r_0 = 1.5$ km, and $\kappa = 400$ km. To reduce GPS and discretization noise, we filter movements shorter than 200 meters, matching the spatial resolution of the processed Shanghai dataset. We do the same for \textit{radius of gyration}, which follows a similar power-law, using the reference values from the same work.

\item \textit{Distance-frequency law}~\cite{Schlapfer2021-pw} is computed as $\rho_i(r,f)=\frac{\mu_i}{(rf)^\eta},$ where $r$ is the distance from the individual's home, $f$ is the visitation frequency over the analysis period, $\mu_i$ represents location attractiveness, and $\eta \approx 2$ is the universal scaling exponent. 

\item \textit{Behavioral mobility profiling}~\cite{licia_sig2020}, locations in individual trajectories are first identified as exploration ($U$) or return ($R$) using the \textit{visitation-frequency-based approach} as in \cite{amichi:hal-03905517}, where $U$ denotes a visit to a previously unseen or rarely visited location and $R$ denotes a revisit to a known location. We compute two features from these sequences: \textit{Intermittency}, which captures how persistently users remain in the same behavioral state before switching to another one ($U\to R$ or $R\to U$), and \textit{Degree of Return}, which measures the overall tendency to revisit familiar locations ($R$) rather than continuously exploring new ones. After min-max normalization, we use Gaussian Mixture Model (GMM) clustering with Silhouette-score optimization to assign users to \textit{Scouters}, \textit{Routiners}, and \textit{Regulars} profiles.

\end{itemize}

\section{Evaluation Setup} 
\label{sec:experimental_setup}

The comparison and validation of simulators' realism across both semantic and topological metrics requires high-quality human mobility datasets. Next, we describe the considered datasets, their processing and sampling procedures applied prior to analysis, as well as the simulator setup used in the realism investigation.

\subsection{Datasets Description and Preprocessing}
\label{ssec:datasets}
We use two spatiotemporal mobility datasets that differ in scale, spatial resolution, and semantic richness.

\noindent\textbf{\parisData{}~\cite{chasse:hal-05365088}:} A non-public GNSS-based anonymized dataset offering a multi-dimensional view of the mobility behavior of 3,337 residents in the Île-de-France region, France, over a continuous 7-day period (collected between October 2022 and May 2023). Volunteers were equipped with the BT-Q1000XT Bluetooth® A-GPS eXtreme Travel Recorder, a GNSS receiver that determines position and velocity using signals from satellite systems. The dataset combines GPS trajectories with validated travel diaries, comprising more than 80,000 trips annotated with visit purposes (e.g., home, work, leisure) and transportation modes (e.g., walk, car, rail). The dataset further includes calibrated population weights designed to correct for sociodemographic and temporal biases, which are incorporated where appropriate in our analyses. All participants provided informed consent for the data collection, which was conducted in accordance with data protection regulations. The start and end locations of each trip are spatially anonymized using the centroid of the corresponding H3 cell (level 10). Since GPS observations are recorded only during movement (i.e., points outside validated trips were removed from the dataset), we infer stationary periods from the trajectory structure and visit annotations.

\begin{itemize}[leftmargin=*]
    \item[-] \textbf{\textit{Preprocessing:}} To reduce noise and spatial fragmentation, we define each visit location as the modal coordinate within a $\pm$10-minute window. Since home/work labels are provided, we consolidate neighboring same-purpose H3 cells within a 2-ring neighborhood to the most frequent centroid. These anonymized centroids are representative spatial anchors, not exact locations. This aggregation increased primary-home coverage from 43.3\% to 75.5\%.
\end{itemize}

\vspace{0.2cm}
\noindent\textbf{Shanghai  \cite{esper_sig24}:}
Cellular Data Record (CDR) dataset collected by a major telecom operator in metropolitan Shanghai, China. The dataset contains hourly mobility records for 58,502 anonymized users over ten consecutive days, totaling more than 9 million observations across 10,396 spatial cells. Unlike CDRs, user locations in this dataset correspond to the centroid of the spatial grid cell nearest to the dominant connected base station during each hour, providing a more stable spatial representation of user mobility. Compared with the \parisData{}, the Shanghai dataset offers substantially larger population coverage but lacks semantic annotations such as activity purposes and transportation modes.

\begin{itemize}[leftmargin=*]
    \item[-] \textbf{\textit{Preprocessing:}} We use a pre-processed version of the Shanghai dataset introduced in prior work~\cite{esper_sig24}, which
addressed common sparsity issues in CDR-like mobility data through temporal completion and spatial normalization~\cite{Chen2019}. Missing records during stable inactivity periods were imputed using dominant home and workplace proxy locations inferred from recurring temporal patterns. Locations were subsequently mapped to a 200m $\times$ 200m OpenStreetMap grid using Scikit-Mobility~\cite{ScikitMobility}, replacing raw coordinates with spatial-cell centroids. To reduce redundancy and remove inactive users, only one observation per user per location was retained within each hourly interval, and users with fewer than 10 active days or 120 total observations were excluded.
\end{itemize}

\subsection{Simulation Setup}
\label{sec:sim_configuration}

\begin{figure}[!htbp]
    \centering
    \begin{subfigure}{0.41\columnwidth}
        \centering
        \includegraphics[width=\linewidth]{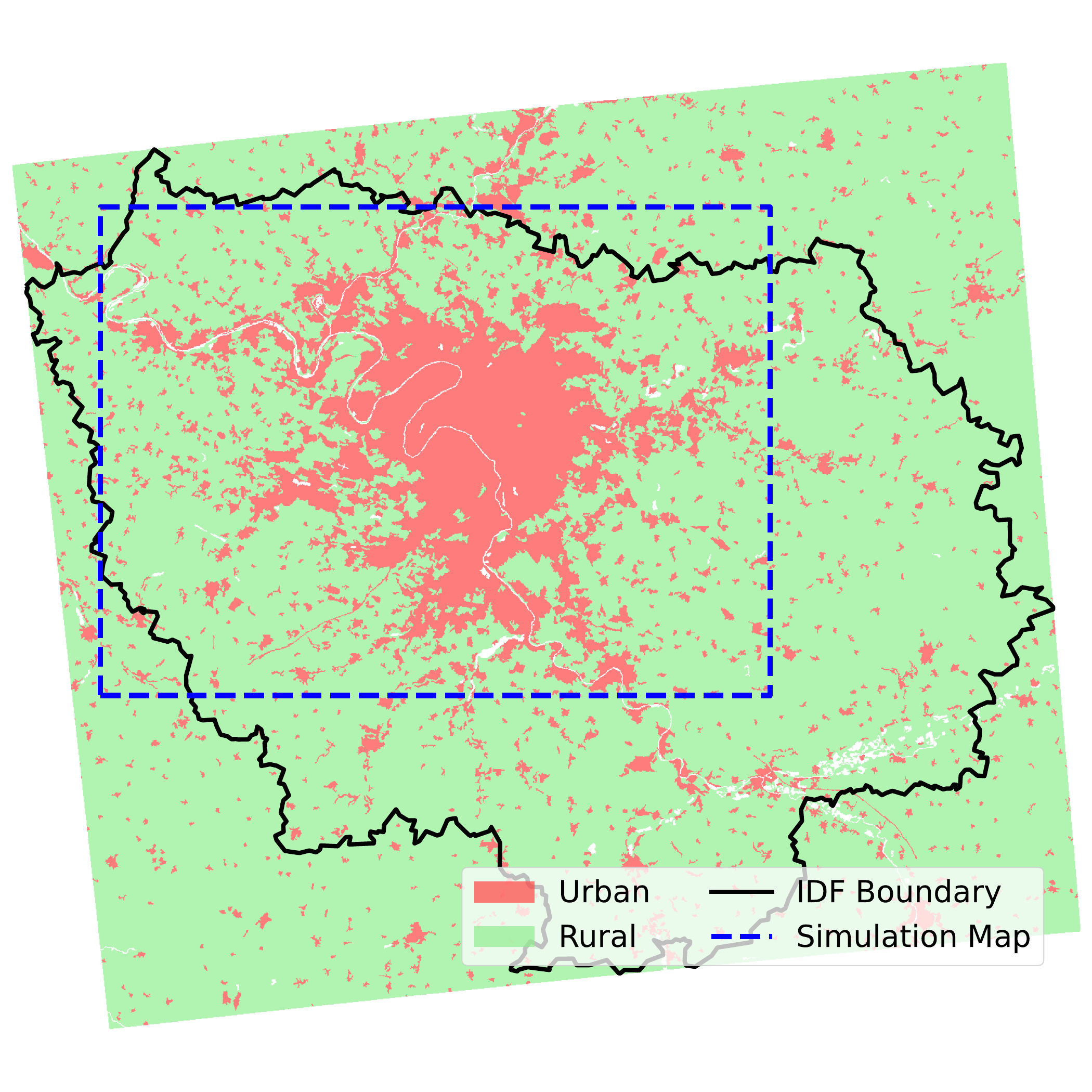}
        \caption{Île-de-France}
        \label{fig:idf_map_1}
    \end{subfigure}
    \hfill
    \begin{subfigure}{0.45\columnwidth}
        \centering
        \includegraphics[height=3.3cm,width=\linewidth]{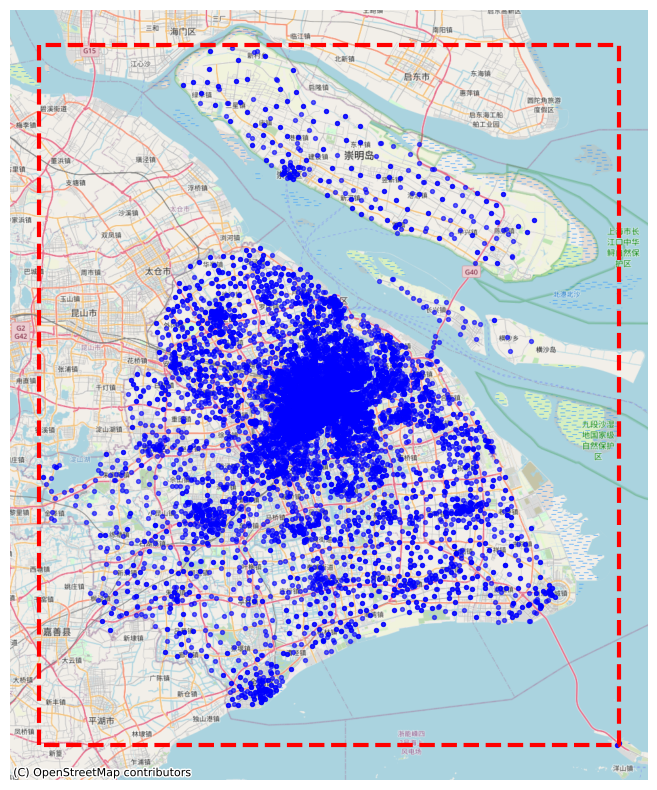}
        \caption{Shanghai}
        \label{fig:shanghai_map}
    \end{subfigure}\vspace{-0.3cm}
    \caption{Simulated bounded regions used in the experiments.}
    \label{fig:simulations_maps_covered}
\end{figure}

To manage computational constraints while preserving representative urban coverage, we define bounded regions for each dataset. For the \parisData{}, instead of simulating the full Île-de-France region, we restrict the environment to a smaller bounding box\footnote{Île-de-France BBOX: (1.6, 48.5, 2.975, 49.16)} covering 61.7\% of the total regional area while encompassing 89.3\% of active urban zones according to Copernicus Land Usage data~\cite{copernicus_land_usage}. For Shanghai, we use a bounding box\footnote{Shanghai BBOX: (120.88, 30.63, 122.06, 31.84)} covering all available mobility records. Figure~\ref{fig:simulations_maps_covered} illustrates both simulation regions.

To match the temporal scope of the empirical datasets, simulations are executed for 7 days in the \parisData{} and 10 days in Shanghai using a 10-minute simulation step. We use the default \textit{AgentSociety} parameter configuration, including the recommended LLM, Qwen-2.5-32B-Instruct model. Results correspond to three independent runs for both \textit{AgentSociety} and \textit{CitySim}.

\subsection{Population Sampling}

Due to the computational cost of LLM-based mobility generation, we downsample both datasets to approximately 500 representative users. This subset size enables the generation of a matching population size of simulated agents and facilitates a direct comparison between real-world and simulated mobility patterns.

In \parisDataShort{}, \textbf{Sample}'s candidate users were restricted to individuals: (i) whose trips remained within the simulation boundaries (cf. \textbf{\S \ref{sec:sim_configuration}}); (ii) with complete 7-day records, including days with no recorded trips, which we treat as non-travel days; and (iii) with stable home and work anchors, defined as representative H3 level-10 cells visited at least five times for home and three times for work. Because this resulted in a subset of 504 users (15\% \textit{Scouters}, 48\% \textit{Regulars}, and 37\% \textit{Routiners}), we retained all of them without additional stratified sampling of profiles. For comparison, the profile distribution in the full dataset is 27\% \textit{Scouters}, 37\% \textit{Regulars}, and 36\% \textit{Routiners}. From this sample, we extract demographic attributes and representative home/work spatial anchors to initialize simulated agents.

In Shanghai, two independent 500-user samples were generated:

\noindent{\textbf{Sample 1:}} We select 500 users with stable spatial anchors, defined as users whose inferred work location remains unchanged across morning and afternoon periods. This restriction is required because both simulators support a single work location per agent. To preserve population-level behavioral diversity, sampling is stratified to maintain the mobility-profile distribution observed in the full dataset (17\% \textit{Scouters}, 56\% \textit{Regulars}, and 27\% \textit{Routiners}).

\noindent{\textbf{Reference sample:}} We draw a second disjoint 500-user sample using the same eligibility criteria to estimate sample-to-sample variability. This sample is not used to initialize agents, but serves as a real-data benchmark for interpreting simulator discrepancies in \textbf{\S \ref{sec:results}}. An analogous GreaterParis reference sample cannot be constructed because too few eligible users remain after filtering.

\section{Mobility Realism Empirical Results}
\label{sec:results}

This section reports the comparison results between simulated and real-world mobility datasets.

\subsection{Spatial Mobility Realism}
\label{sec:results_spatial}

\greybox{\textit{LLM-based agents capture the main qualitative mobility-law patterns observed in the literature and empirical datasets, but fail to accurately capture individual-level spatial mobility behavior.}} \vspace{-0.3cm}

\begin{figure}[htb]

    \includegraphics[width=1\linewidth]{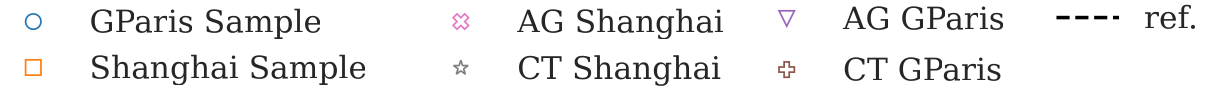}
    \vspace{0.3cm} % Um pequeno espaço entre a legenda e os gráficos
  \centering
  \subfloat[Travel Distance ($\Delta r$)]{
    \includegraphics[width=0.22\textwidth, keepaspectratio]{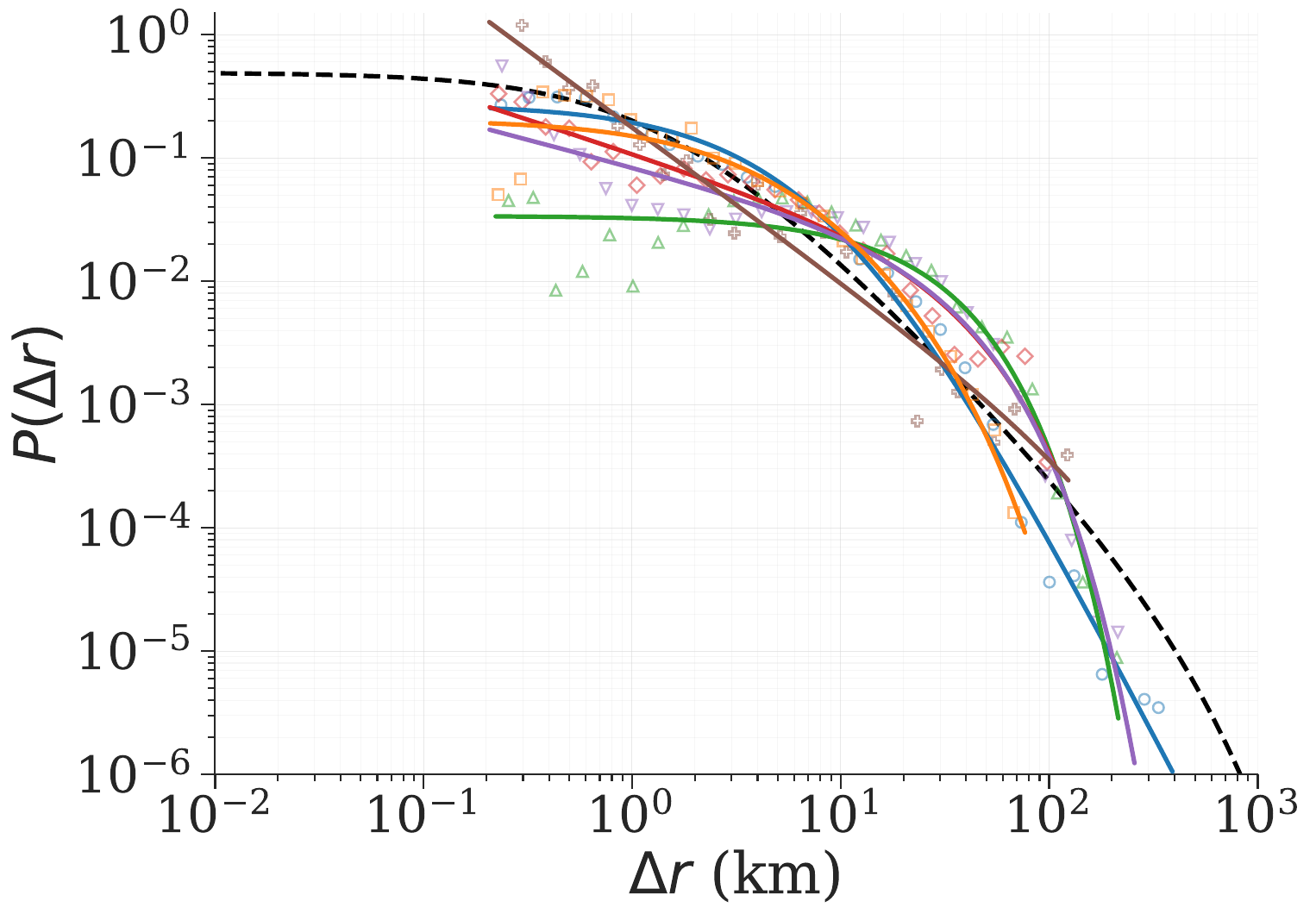}
  }
  \hfill
  \subfloat[\textit{Radius of gyration} ($r_g$)]{
    \includegraphics[width=0.22\textwidth, keepaspectratio]{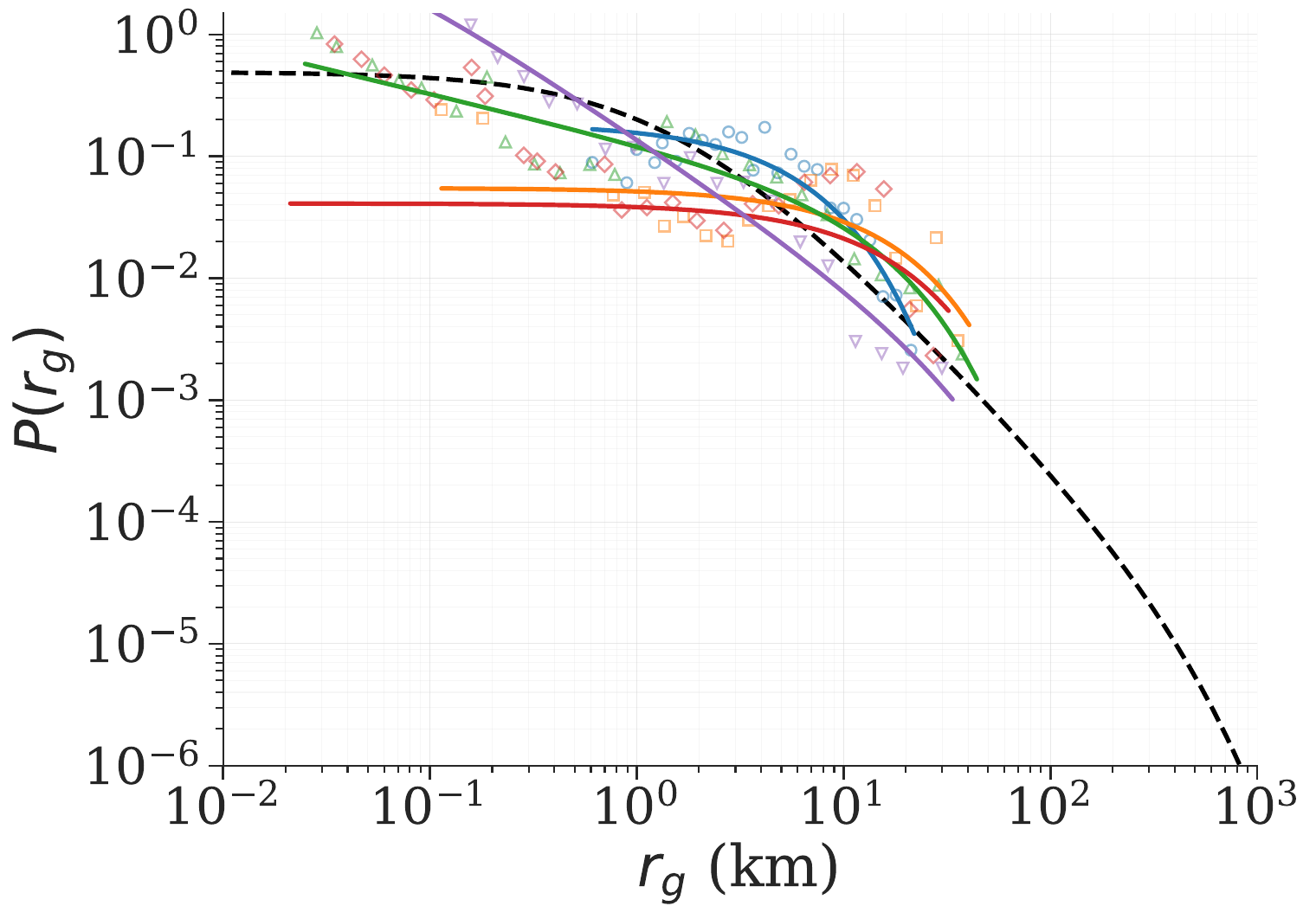}
  }\vspace{-0.2cm}
  \hfill
  \subfloat[Distance-Frequency Law]{
    \includegraphics[width=0.22\textwidth, keepaspectratio]{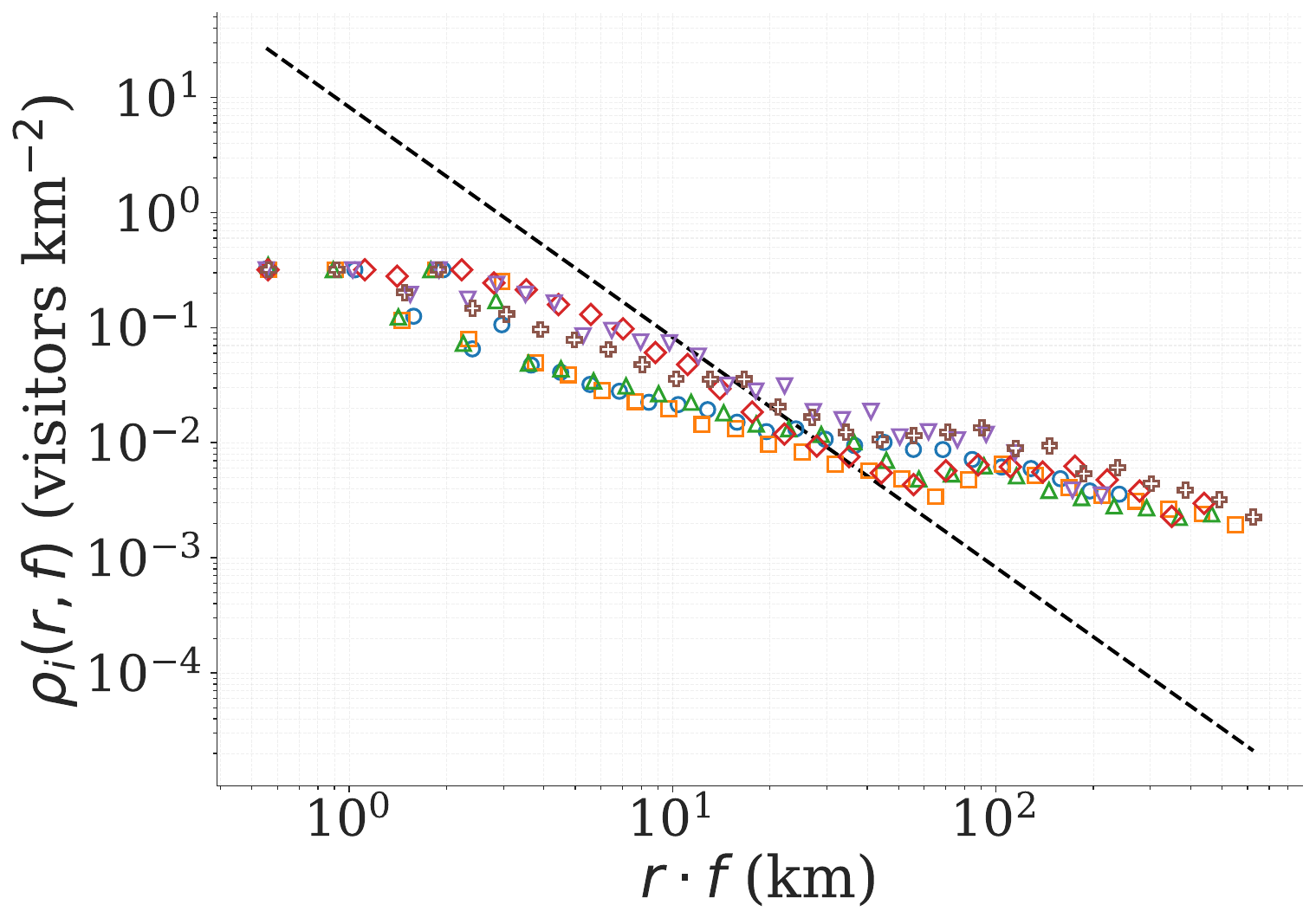}
  }
  \hfill
  \subfloat[Daily Visits Law]{
    \includegraphics[width=0.22\textwidth, keepaspectratio]{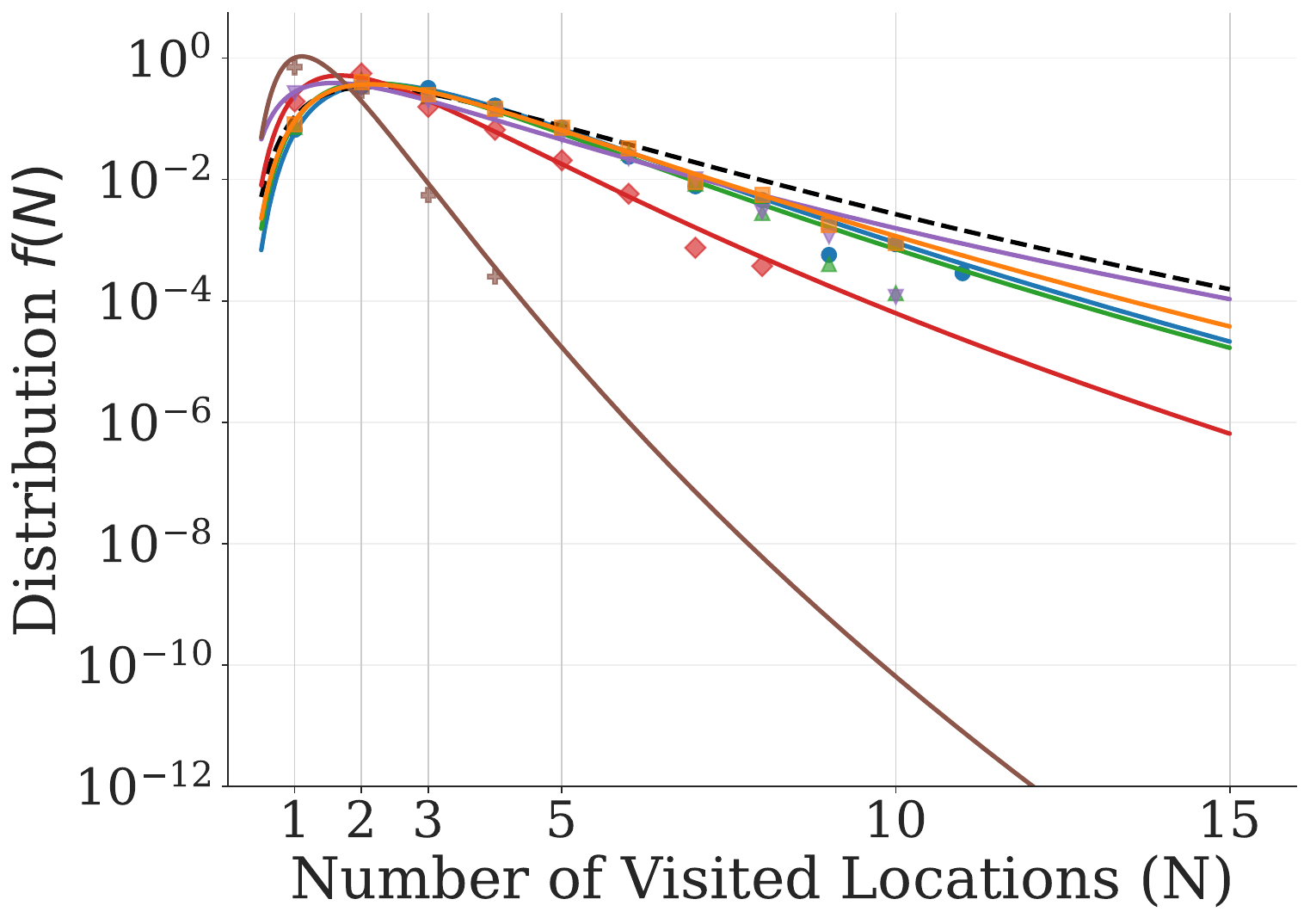}
  }\vspace{-0.3cm}
  \caption{Comparison of empirical mobility-law distributions between real and simulated trajectories across \parisDataShort{} and Shanghai. AG refers to \textit{AgentSociety} and CT to \textit{CitySim}.} 
  \label{fig:mobility_laws_comparision}
\end{figure}

Fig.~\ref{fig:mobility_laws_comparision} compares the empirical mobility-law distributions observed in the real sampled datasets (denoted as GParis/Shanghai Sample) with those generated by \textit{AgentSociety} and \textit{CitySim} (denoted as AG/CT GParis or AG/CT Shanghai). The corresponding theoretical reference distributions reported in the literature are shown as dotted lines and labeled ``ref'' in the figure.

Overall, both simulators capture the main qualitative mobility-law patterns observed in literature and sampled datasets; the generated trajectories exhibit: (i) \textit{travel-distance} distributions consistent with a truncated power law, characterized by a predominance of short trips and progressively fewer long-distance movements~\cite{Gonzalez2008}; (ii) radius-of-gyration distributions indicating spatially bounded daily mobility, with most individuals exhibiting confined mobility within a few kilometers of their primary activity locations~\cite{Gonzalez2008}; (iii) a distance--frequency scaling relationship in which visitation frequency decreases with distance from home, reflecting mobility patterns dominated by frequent short-range visits and occasional long-range trips~\cite{Schlapfer2021-pw}; and (iv) a daily number of visited locations that approximately follows a log-normal distribution~\cite{Schneider2013}.

\begin{table}[tbh]
\centering
\scriptsize
\caption{Comparison of real vs simulated distribution of spatial mobility metrics.}
\vspace{-0.3cm}
\label{tab:spatial_results}
\begin{tabular}{llccc}
\toprule
\textbf{Dataset} & \textbf{Source} & \textbf{\textit{$\Delta r$}} ($W_1$) & \textbf{\textit{$r_g$}} ($W_1$) \\
\midrule

\parisDataShort{} & \textit{AgentSociety} & $14.83 \pm 0.35$ & $7.29 \pm 0.28$  \\

\parisDataShort{} & \textit{CitySim} & $7.53 \pm 1.67$ & $3.47 \pm 0.23$  \\

Shanghai & \textit{AgentSociety} & $8.73 \pm 0.48$ & $4.30 \pm 0.67$  \\

Shanghai & \textit{CitySim} & $3.97 \pm 0.11$ & $4.86 \pm 0.09$ \\

\midrule
Shanghai & Reference sample & $0.48$ & $1.01$\\

\bottomrule
\end{tabular}
\end{table}

Tables~\ref{tab:spatial_results} and \ref{tab:spatial_fidelity_resolutions} summarize the spatial realism results across datasets and simulators.
Table~\ref{tab:spatial_results} compares the \textit{travel-distance} ($\Delta r$) and \textit{radius of gyration} (\textit{$r_g$}) distributions of the real and simulated datasets using the Wasserstein distance ($W_1$), where lower values indicate greater similarity between real and simulated mobility patterns.
Additionally, Table~\ref{tab:spatial_fidelity_resolutions} (i) reports $W_1$ values comparing simulated against empirical mobility patterns using \textit{Spatio-Temporal Visit Distribution} (STVD) at different spatial resolutions and (ii) evaluates the similarity between real and simulated OD matrices using the Common Part of Commuters (CPC) metric. Higher CPC values indicate greater similarity between mobility flows. Results obtained using the complete dataset populations are reported in Appendix \ref{sec:appx_spatial_cdr_population}. Relative to the Shanghai reference sample, simulator errors are much larger,  indicating deviations beyond sampling variability.

As shown, datasets generated by \textit{AgentSociety} exhibit substantial spatial discrepancies in both  \parisData{} and Shanghai scenarios. These discrepancies are reflected in the \textit{$\Delta r$} and \textit{$r_g$} distributions, which consistently yield larger $W_1$ %values 
than those obtained with \textit{CitySim}. 
Similar discrepancies are observed for the STVD metric in both simulators, indicating partial reproduction of visit spatial distributions. At the flow level, OD-matrix similarity remains weak across all scenarios, as reflected by the low CPC values, suggesting that neither \textit{AgentSociety} nor \textit{CitySim} fully captures realistic large-scale mobility flows between locations. These results suggest that the generated trajectories capture some high-level behavioral structure while failing to reproduce these dynamics.
 
 \greybox{\textit{Spatial realism limitations are more pronounced in \textit{AgentSociety} outputs. Analysis of the generation pipeline suggests that %these 
 mismatches are partially associated with instability in POI selection and destination reasoning, leading to trajectories influenced by local map-density artifacts rather than coherent large-scale mobility patterns.}} \vspace{-0.3cm}

\textit{CitySim} partially mitigates these spatial inconsistencies. In the \parisData{}, \textit{$\Delta r$} error decreases from $14.83$ km (in \textit{AgentSociety} case) to $7.53$ km, while \textit{radius of gyration} error decreases from $7.29$ km to $3.47$ km. In Shanghai, \textit{CitySim} further reduces \textit{$\Delta r$} error to $3.97$ km and \textit{radius of gyration} error to $4.86$ km. Nevertheless, OD-matrix agreement remains weak, particularly in the \parisData{} (\textit{CPC} H8 $=0.138\pm0.018$). 

 \greybox{\textit{While CitySim improves destination and spatial selection relative to \textit{AgentSociety}, these enhancements alone do not yield realistic large-scale mobility patterns.}}\vspace{-0.3cm}

\begin{figure}[]
    \centering
    \begin{subfigure}{0.48\columnwidth}
        \centering
        \includegraphics[height=3.6cm, width=\linewidth]{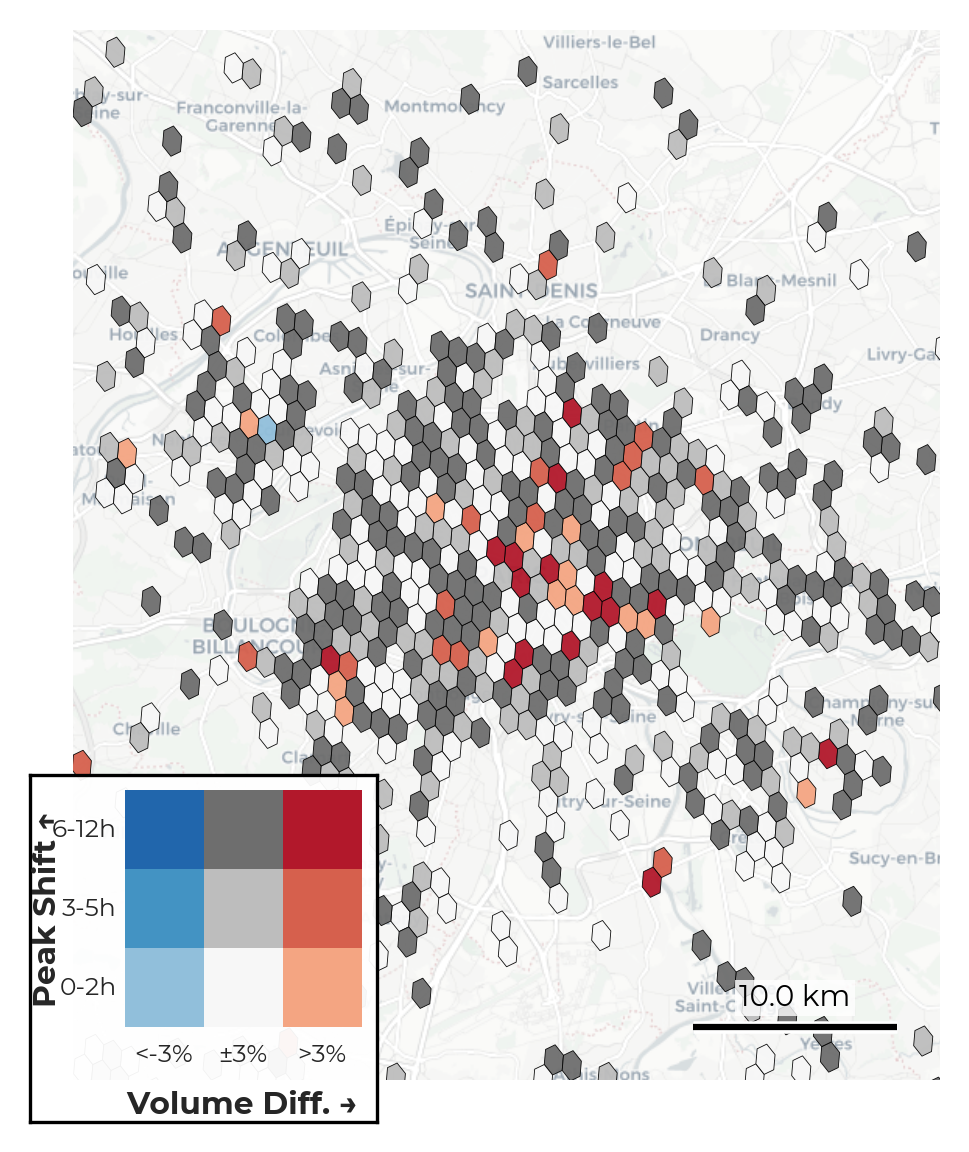}
        \caption{\parisDataShort{}}
        \label{fig:idf_map_stvd}
    \end{subfigure}
    \hfill
    \begin{subfigure}{0.48\columnwidth}
        \centering
        \includegraphics[height=3.6cm,width=\linewidth]{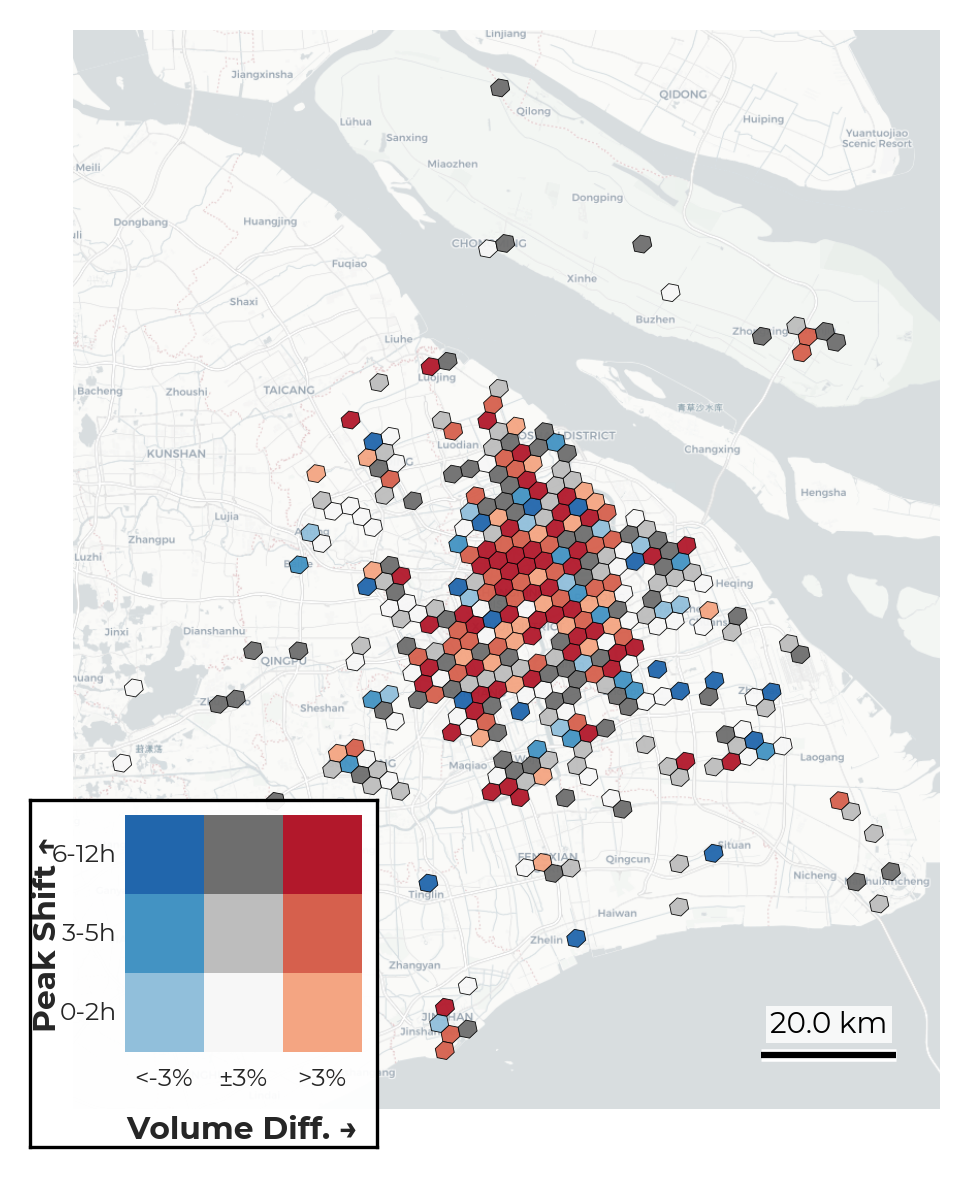}
        \caption{Shanghai}
        \label{fig:shanghai_map_stvd}
    \end{subfigure}\vspace{-0.3cm}
    \caption{\textit{Spatio-Temporal Visit Distribution} (\textit{STVD}) comparing real and simulated mobility patterns.}
    \label{fig:stvd}
\end{figure}

To further examine the spatiotemporal fidelity results reported in Table~\ref{tab:spatial_fidelity_resolutions}, we analyze the STVD in greater detail. Figure~\ref{fig:stvd} presents hexagonal bivariate maps that jointly visualize visit-volume differences and temporal displacement of peak activity (Peak Shift) between simulated and empirical mobility patterns. For clarity, we show only the \textit{AgentSociety} simulations, as
\textit{AgentSociety} achieves lower STVD $W_1$ values than \textit{CitySim}, despite \textit{CitySim}'s better agreement for individual-level \textit{$\Delta r$} and \textit{$r_g$} distributions. At H3 resolution 8, the \parisData{} STVD $W_1$ equals $607.69 \pm 5.88$ for \textit{AgentSociety} and $748.71 \pm 23.00$ for\textit{CitySim}. For the same resolution, the Shanghai STVD $W_1$ is also lower for \textit{AgentSociety}. Results suggest that the simpler gravity-based destination selection used in \textit{AgentSociety} outperforms \textit{CitySim}'s LLM-driven POI-selection strategy in reproducing the most frequently visited urban areas.

\greybox{\textit{Increasing POI-selection complexity does not necessarily improve the reproduction of highly visited urban areas.}}\vspace{-0.3cm}

In \parisData{} (Fig.~\ref{fig:idf_map_stvd}), the simulation captures the broad spatial distribution of user activity but exhibits noticeable temporal discrepancies. Agents tend to visit the correct regions, yet peak activity is often shifted by 3 to 12 hours, suggesting that the model predicts where urban activity occurs more accurately than when it occurs. Aggregate visit volumes are also slightly underestimated in central Paris. In contrast, Shanghai (Fig.~\ref{fig:shanghai_map_stvd}) exhibits stronger spatiotemporal divergence in the urban core, characterized by dense clusters of visit-volume and temporal errors (red hexagons). This divergence may stem from dataset-specific limitations, such as sparse contextual information or challenges in extracting POI metadata, though peripheral areas align more closely with the empirical baseline.
Recall that STVD similarity is measured using an approximate Wasserstein distance in which a temporal difference of 10 minutes incurs the same penalty as a spatial displacement of 100 m (cf.  Table~\ref{tab:spatial_fidelity_resolutions}).

\begin{table}[tbh]
\centering
\scriptsize
\caption{Spatio-Temporal realism metrics 
comparing datasets' \textit{STVD} %distances 
and \textit{OD matrix} across resolutions. }
\vspace{-0.3cm}
\label{tab:spatial_fidelity_resolutions}
\begin{tabular}{llccc}
\toprule
\textbf{Dataset} & \textbf{Source} & \textbf{Resolution} & \textbf{\textit{STVD}} ($W_1$) & \textbf{\textit{OD Matrix}} (\textit{CPC}) \\
\midrule
\multirow{6}{*}{\parisDataShort{}} & \multirow{3}{*}{\textit{AgentSociety}} & H7 & $505.83 \pm 7.20$ & $0.206 \pm 0.004$ \\
 & & H8 & $607.69 \pm 5.88$ & $0.088 \pm 0.002$ \\
 & & H9 & $694.77 \pm 2.66$ & $0.027 \pm 0.001$ \\
\cmidrule(lr){2-5}
 & \multirow{3}{*}{\textit{CitySim}} & H7 & $628.60 \pm 6.07$ & $0.270 \pm 0.026$ \\
 & & H8 & $748.71 \pm 23.00$ & $0.138 \pm 0.018$ \\
 & & H9 & $844.26 \pm 22.18$ & $0.038 \pm 0.006$ \\
\midrule
\multirow{9}{*}{Shanghai} & \multirow{3}{*}{\textit{AgentSociety}} & H7 & $111.16 \pm 42.77$ & $0.135 \pm 0.019$ \\
 & & H8 & $113.88 \pm 39.04$ & $0.046 \pm 0.008$ \\
 & & H9 & $145.02 \pm 60.10$ & $0.010 \pm 0.003$ \\
\cmidrule(lr){2-5}
 & \multirow{3}{*}{\textit{CitySim}} & H7 & $422.11 \pm 17.32$ & $0.054 \pm 0.013$ \\
 & & H8 & $432.55 \pm 17.11$ & $0.042 \pm 0.013$ \\
 & & H9 & $449.21 \pm 31.84$ & $0.011 \pm 0.005$ \\
\cmidrule(lr){2-5}
 & \multirow{3}{*}{Reference sample} & H7 & $21.11$ & $0.297$ \\
 & & H8 & $6.29$ & $0.092$ \\
 & & H9 & $35.02$ & $0.045$ \\
\bottomrule
\end{tabular}
\end{table}

\subsection{Temporal Mobility Dynamics}

\begin{figure}[h]
    \centering
    \includegraphics[width=1\linewidth]{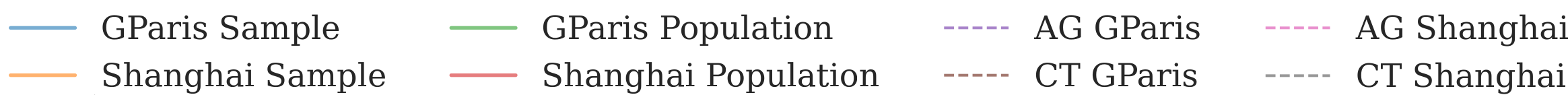}
    \subfloat[Trip Duration]{
        \includegraphics[height=2.5cm,width=0.15\textwidth]{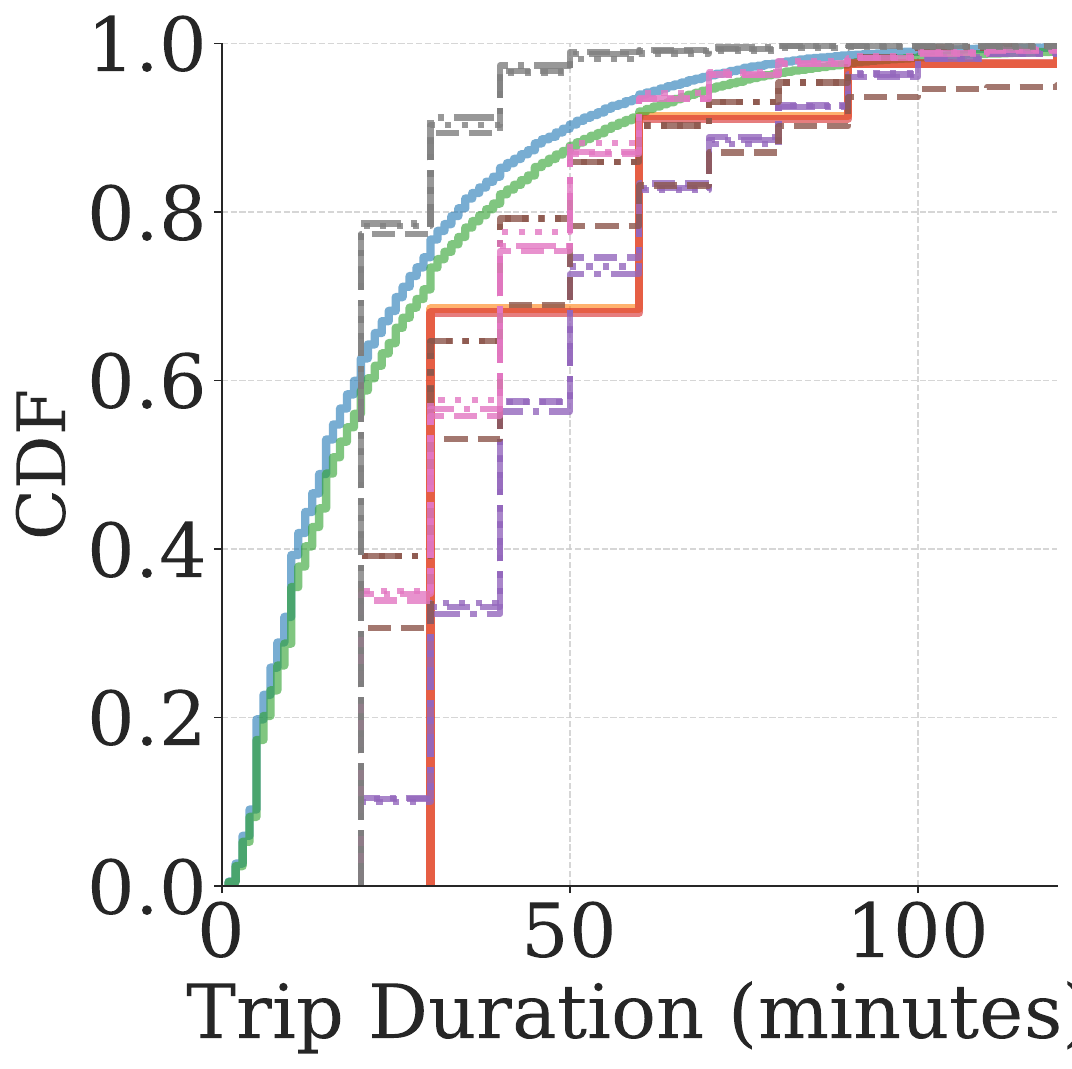}
    }
    \subfloat[Dwell Time]{
        \includegraphics[height=2.5cm,width=0.15\textwidth]{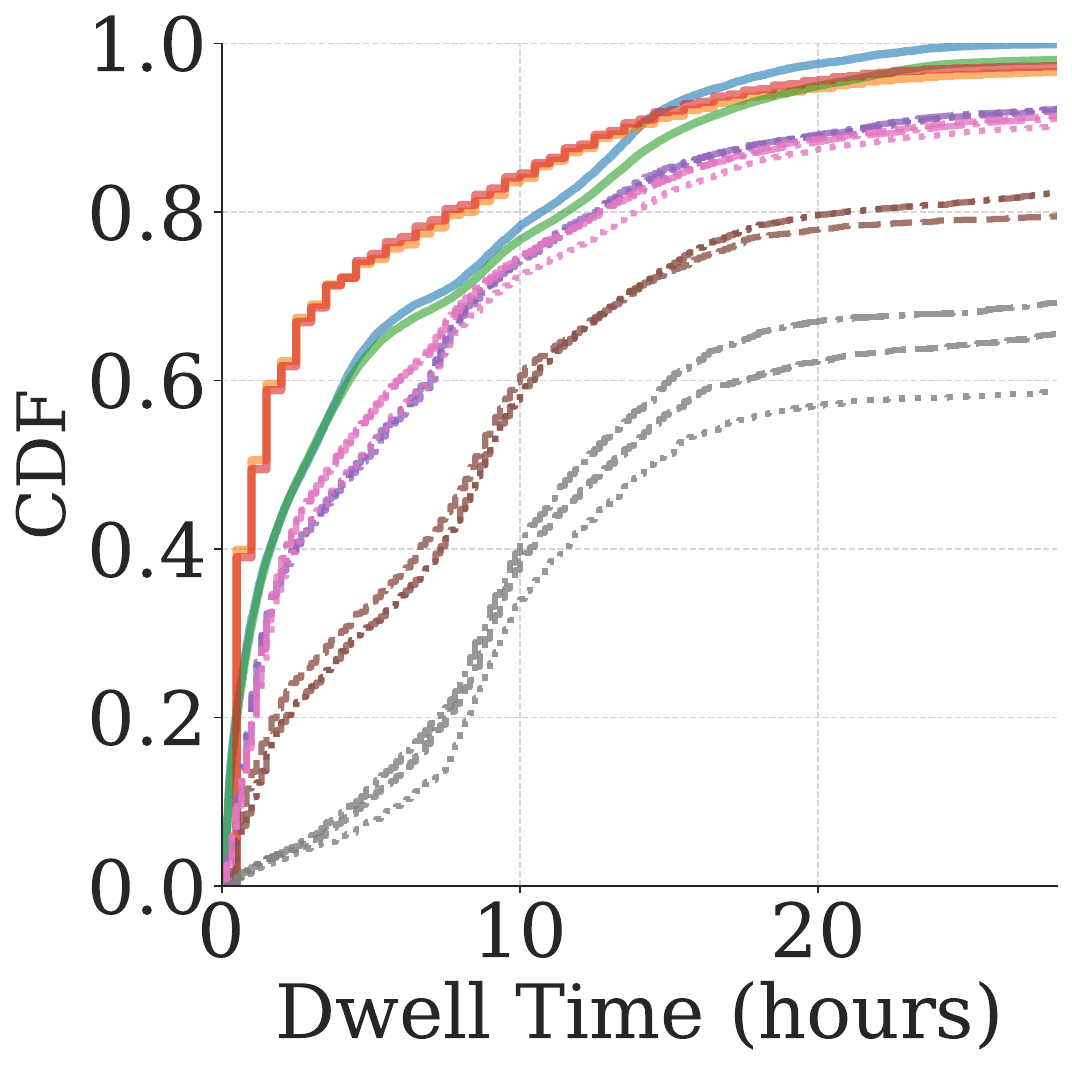}
    }
    \subfloat[Visit Frequency]{
        \includegraphics[height=2.5cm,width=0.15\textwidth]{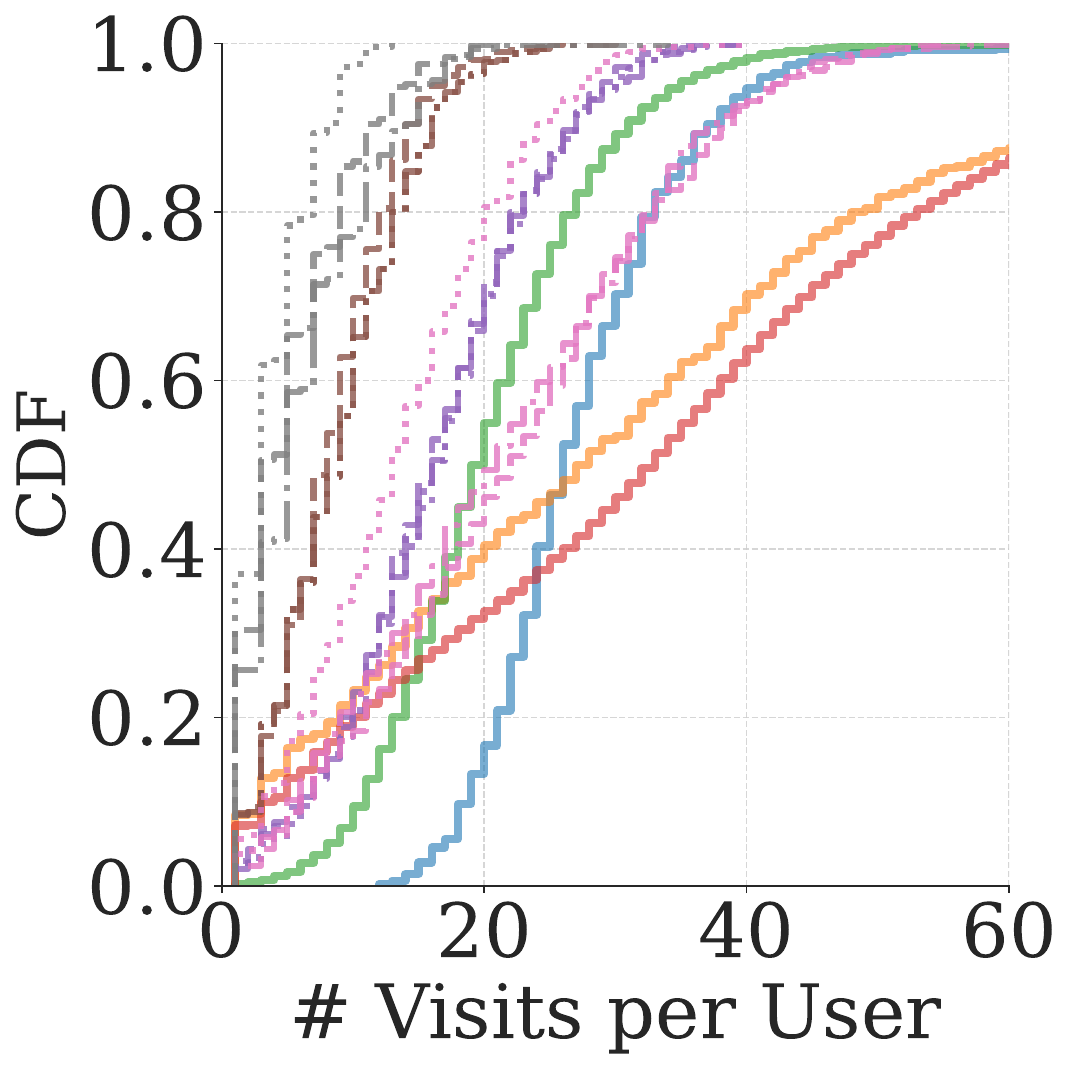}
    }\vspace{-0.3cm}
    \caption{Comparison of temporal distributions between real and simulated trajectories across \parisDataShort{} and Shanghai.}
    \label{fig:temporal_rhythms}
\end{figure}

\greybox{\textit{Spatial deviations propagate into unrealistic temporal dynamics.}}\vspace{-0.3cm}
Fig.~\ref{fig:temporal_rhythms} compares the temporal distributions observed in the empirical datasets with those generated by \textit{AgentSociety} and \textit{CitySim}, while Table~\ref{tab:temporal_results} summarizes the main temporal realism metrics across datasets and simulators. Since destination selection and travel distances jointly influence travel and waiting times, deviations in spatial behavior are expected to impact daily temporal rhythms. 

\textit{AgentSociety} exhibits substantial discrepancies in \textit{trip duration}, \textit{dwell time}, and \textit{visitation frequency} across both datasets. In the \parisData{}, \textit{trip duration} error reaches $W_1=24.41\pm0.24$ minutes, while \textit{dwell time} shows a large mismatch ($W_1= 3.96\pm0.12$ hours). Similar patterns appear in Shanghai with lower absolute errors, likely due to the dataset’s coarser temporal resolution. Regarding \textit{visitation frequency}, which empirically follows a log-normal distribution, \textit{AgentSociety} captures the overall shape of the distribution but consistently underestimates \textit{daily visits}. Over the multi-day simulation, this daily underestimation accumulates, leading to a substantial deficit in total visits across both scenarios.

\begin{table}[t]
\centering
\scriptsize
\caption{Temporal mobility realism metrics comparing real and simulated trajectories.}
\vspace{-0.3cm}
\label{tab:temporal_results}
\begin{tabular}{llccc}
\toprule
\textbf{Dataset} & \textbf{Source} & \textbf{\textit{TD.} (min)} ($W_1$) & \textbf{\textit{DT.} (h)} ($W_1$) & \textbf{ \textit{Vf.}} ($W_1$) \\
\midrule

\parisDataShort{} & \textit{AgentSociety} & $24.41 \pm 0.24$ & $3.96 \pm 0.12$ & $9.63 \pm 0.19$ \\

\parisDataShort{} & \textit{CitySim} & $24.32 \pm 18.93$ & $30.24 \pm 29.63$ & $19.46 \pm 3.62$ \\

Shanghai & \textit{AgentSociety} & $9.30 \pm 0.24$ & $4.79 \pm 0.37$ & $12.31 \pm 4.19$ \\

Shanghai & \textit{CitySim} & $19.30 \pm 0.27$ & $36.39 \pm 4.91$ & $24.95 \pm 0.65$ \\
\midrule
Shanghai & Ref. sample & $0.29$ & $0.26$ & $2.07$ \\

\bottomrule
\end{tabular}
\end{table}

Although \textit{CitySim} improves spatial realism metrics such as \textit{$\Delta r$} and \textit{$r_g$} (cf. \S~\ref{sec:results_spatial}), Table~\ref{tab:temporal_results} shows that these gains are accompanied by a degradation in temporal realism. The inclusion of goals/hobbies (e.g., looking for bird spotting POI categories instead of parks) and more complex LLM prompts for POI selection leads to longer decision times and increased residence time at home locations, reducing overall mobility activity. 

\greybox{\textit{Mobility realism requires jointly reproducing spatial visitation patterns and temporal activity schedules; improving spatial realism alone may degrade temporal consistency.}}\vspace{-0.3cm}

Compared with the Shanghai reference sample, whose temporal discrepancies are small ($W_1=0.29$ minutes for \textit{trip duration}, $W_1=0.26$ hours for
\textit{dwell time}, and $W_1=2.07$ for \textit{visitation frequency}), both
simulators exhibit substantially larger errors. 

\greybox{\textit{Temporal discrepancies reflect systematic biases in the simulated mobility dynamics rather than finite-sample variability.}}

\begin{figure}[htb]
    \centering
    \begin{subfigure}{0.49\columnwidth}
        \centering
        \includegraphics[width=0.99\linewidth]{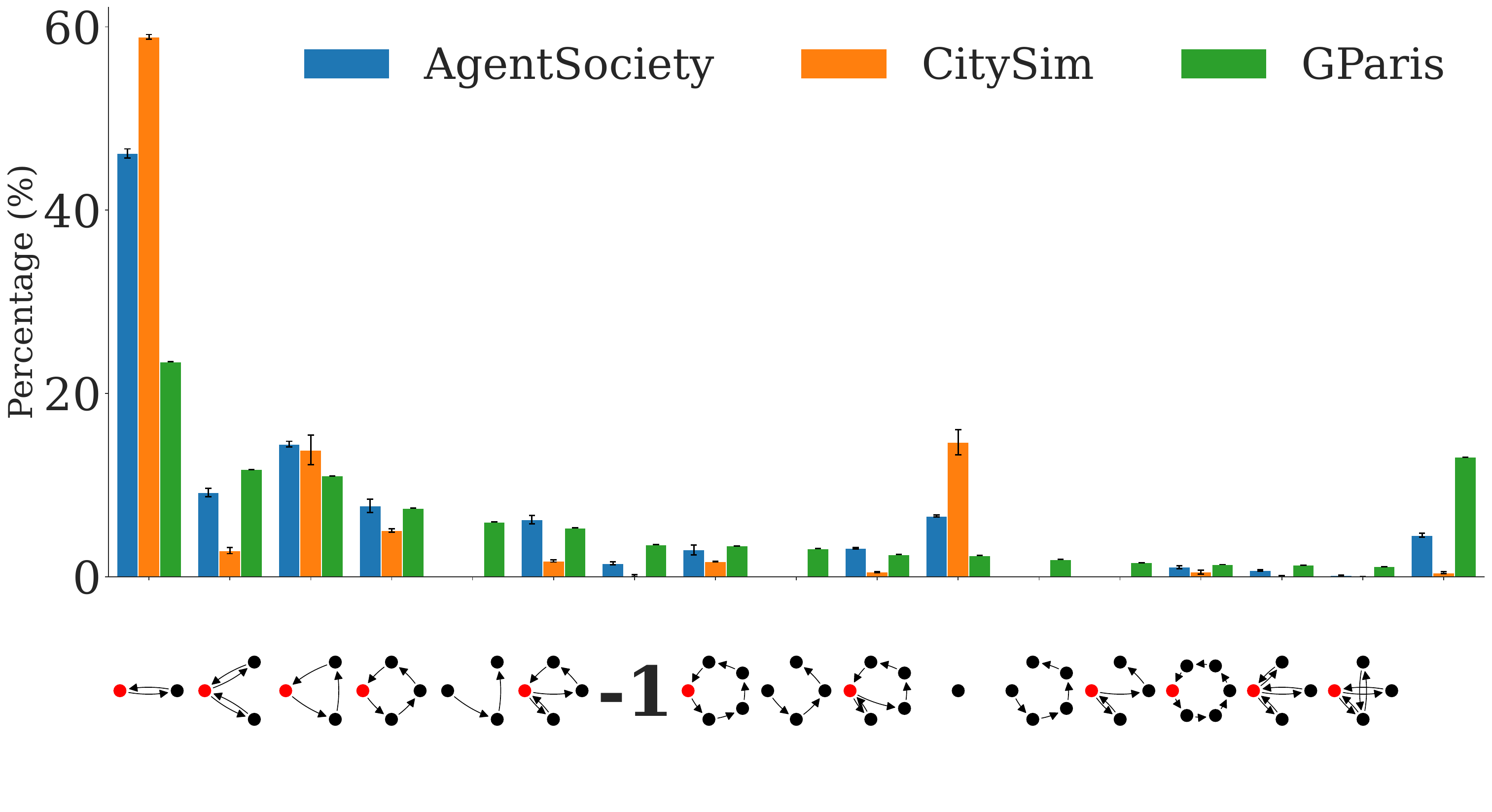}
        \caption{\parisDataShort{} Motifs}
        \label{fig:netmob_motifs}
    \end{subfigure}
    \hfill
    \begin{subfigure}{0.49\columnwidth}
        \centering
        \includegraphics[width=0.99\linewidth]{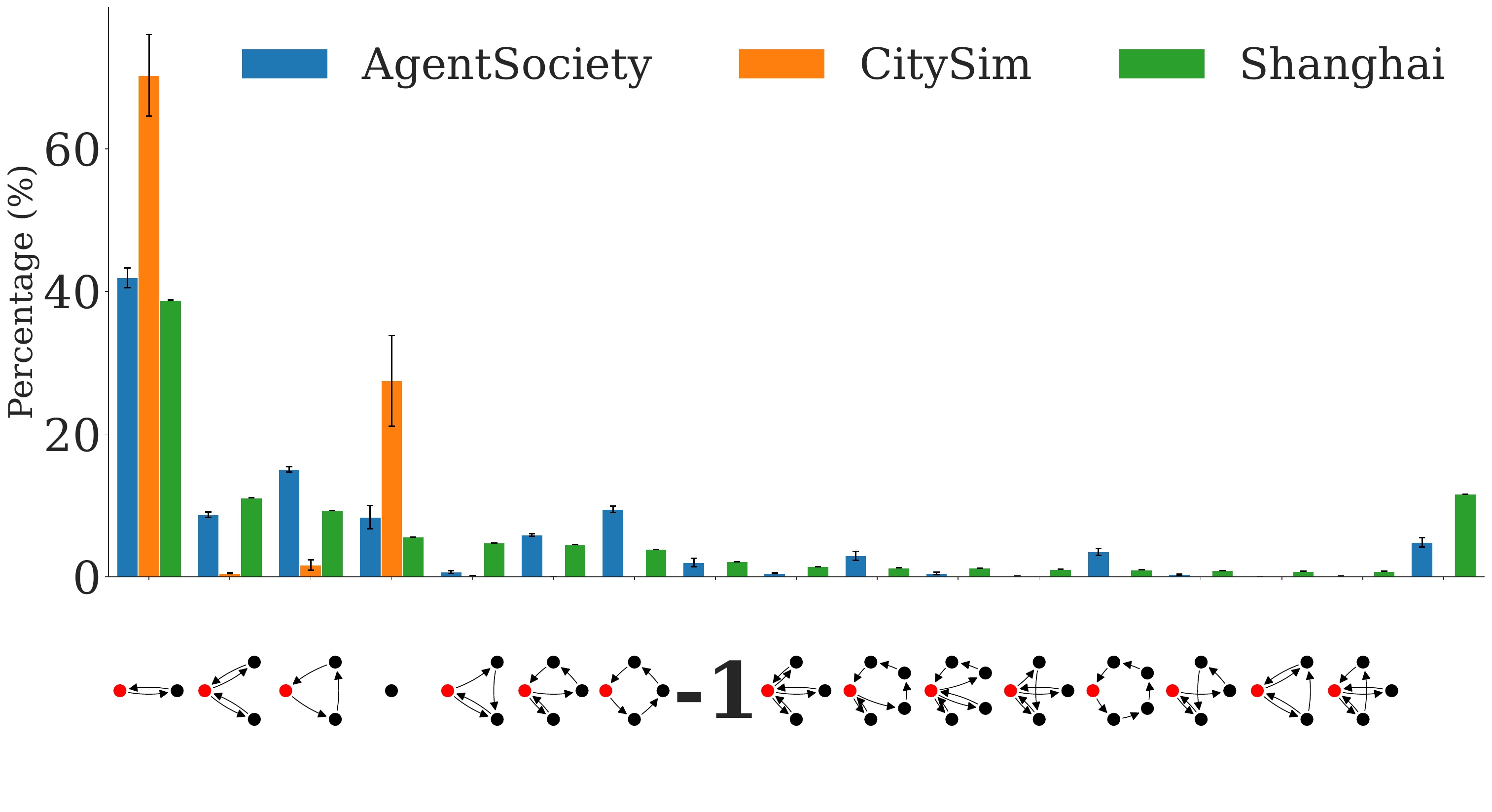}
        \caption{Shanghai Motifs}
        \label{fig:shanghai_motifs}
    \end{subfigure}
    \hfill \vspace{-0.3cm}
    \caption{Comparison of the most frequent daily \textit{mobility motifs} observed in real and simulated trajectories.}
    \label{fig:motifs}
\end{figure}

\subsection{Mobility Motifs}
\greybox{\textit{Simulated trajectories reproduce dominant routine motifs but fail to capture the full diversity of empirical daily mobility structures.}}\vspace{-0.3cm}

Fig.~\ref{fig:motifs} shows the 16 most frequent \textit{mobility motifs}~\cite{Schneider2013} as well as their distributions observed in all  
datasets, allowing comparison between empirical and simulated outputs.
Motif label $-1$ denotes motifs with more than six nodes, while the unlabeled bar aggregates motifs outside the top 16 motifs observed in the \parisDataShort{} and Shanghai datasets. Table~\ref{tab:motif_results} reports the \textit{JSD} between real and simulated motif distributions.  The columns indicate the reference distribution against which each \textit{Source} is compared: the corresponding 500-user empirical sample (\textit{Sample}), the full empirical dataset (\textit{Population}), or the canonical motif distribution reported in~\cite{Schneider2013} (\textit{Literature}). The rows quantify how far the main 500-user empirical sample deviates from the full population and literature references, while the Shanghai reference-sample row quantifies the variability between two disjoint empirical Shanghai samples of the same size.

\begin{table}[tbh]
\centering
\scriptsize
\caption{\textit{JSD} for
\textit{mobility motif} distributions. \textit{Sample}, \textit{Population},
and \textit{Literature} denote the reference distribution used in \textit{Source} comparisons.} 
\vspace{-0.3cm}
\label{tab:motif_results}
\begin{tabular}{llccc}
\toprule
\textbf{Dataset} & \textbf{Source} & \textbf{Sample} & \textbf{Population} & \textbf{Literature} \\
\midrule
\parisDataShort{} & \textit{AgentSociety} & $0.1186 \pm 0.0026$ & $0.1290 \pm 0.0013$ & $0.0463 \pm 0.0030$ \\
\parisDataShort{} & \textit{CitySim}      & $0.3272 \pm 0.0975$ & $0.2845 \pm 0.1193$ & $0.2034 \pm 0.0789$ \\
Shanghai & \textit{AgentSociety} & $0.0612 \pm 0.0027$ & $0.1084 \pm 0.0109$ & $0.0406 \pm 0.0036$ \\
Shanghai & \textit{CitySim}      & $0.3298 \pm 0.0206$ & $0.4489 \pm 0.0199$ & $0.3447 \pm 0.0109$ \\
\midrule
\parisDataShort{}  & Empirical sample    & \textit{n/a} & $0.006$ & $0.05$\\
Shanghai & Empirical sample      & \textit{n/a} & $0.05$ & $0.04$ \\
Shanghai & Reference sample     & $0.0045$ & $0.0620$ & $0.04$  \\
\bottomrule
\end{tabular}
\end{table}

Across both datasets, the simulations are strongly dominated by simple two-node motifs, indicating that agents frequently alternate between a small number of anchors, such as home and work. While this captures part of the repetitive structure observed in real mobility, many empirical daily routines remain unmatched.

In the \parisData{}, mismatch is partly associated with open-ended daily trajectories in which individuals start and end the day in different locations, producing unclosed motif loops. In Shanghai, mismatch is more strongly driven by the higher spatial granularity and density of recorded locations, which frequently generate complex motifs exceeding six nodes and therefore fall outside the evaluated motif space. Finally, the Shanghai reference-sample JSD is much smaller than the simulators' JSDs in the \textit{Sample} column, indicating that motif mismatches are substantially
larger than what would be expected from sampling variability alone.

\textit{Predictability} gives the same pattern a useful second angle. In \parisDataShort{}, empirical trajectories have mean \textit{predictability} $0.472$, while \textit{AgentSociety} and \textit{CitySim} increase this value to $0.539\pm0.007$ and $0.598\pm0.114$, respectively, consistent with an overproduction of simple routine structures. In Shanghai, the empirical value is higher ($0.666$); \textit{AgentSociety} underestimates it ($0.501\pm0.014$), whereas \textit{CitySim} is closer but slightly higher ($0.689\pm0.005$). 
\greybox{\textit{Similar motif-level errors can arise from different degrees of \textit{regularity} in the generated trajectories.}} \vspace{-0.2cm}

Despite these limitations, the middle of the ranked motif distribution exhibits partially similar variability between empirical and simulated trajectories, suggesting that LLM-based agents capture some coarse routine regularities while failing to reproduce the full topological diversity of real-world mobility behavior.

\begin{figure}[htb]
    \centering
    \includegraphics[width=0.9\linewidth]{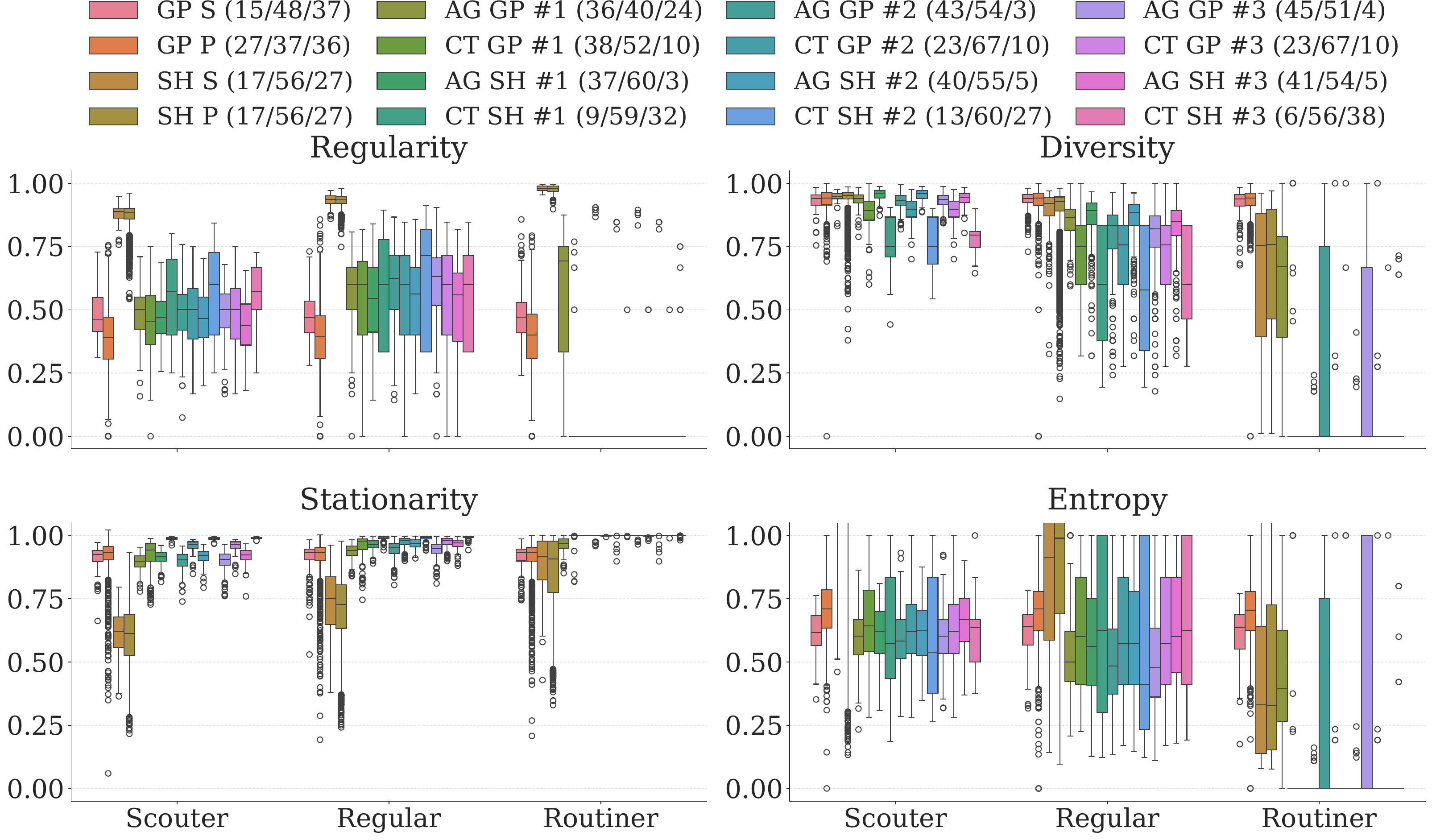}\vspace{-0.3cm}
    \caption{Behavioral mobility metrics and \textit{profile} distributions across empirical and simulated trajectories.}
    \label{fig:15_profiling_mobility_metrics}
\end{figure}

\subsection{Behavioral Mobility Profiles}

\greybox{\textit{Simulated trajectories partially reproduce the expected behavioral mobility profile structure with routine and exploration dynamics.} } \vspace{-0.3cm}
Fig.~\ref{fig:15_profiling_mobility_metrics} summarizes the behavioral mobility metrics~--\textit{regularity}, \textit{diversity}, \textit{stationarity}, and \textit{entropy}~-- associated with the profile distributions in parentheses in the legend (\textit{Scouters}/\textit{Regulars}/\textit{Routiners}) across empirical and simulated datasets. S, P, GP, and SH refer to Sample, Population, GreaterParis, and Shanghai, respectively.

Overall, the simulated \textit{profiles} broadly align with mobility-profiling literature: \textit{Routiners} exhibit higher \textit{regularity} and \textit{stationarity} with lower \textit{diversity} and \textit{entropy}, while \textit{Scouters} exhibit the opposite behavior. This suggests that LLM-based agents capture part of the exploration-return dynamics underlying human mobility. The \parisData{} profile-distribution \textit{JSD} remains relatively low for \textit{AgentSociety} ($0.13 \pm 0.08$), while \textit{CitySim} is slightly lower ($0.09 \pm 0.01$).

Some Shanghai simulation runs exhibit extreme outliers with unusually high \textit{diversity} and \textit{entropy}. These anomalies appear consistent with instability in POI-selection behavior, where agents repeatedly search for unavailable or overly specific destination types, producing erratic movement patterns that artificially increase exploratory behavior and shift agents toward the \textit{Scouter} profile. Across most runs, a small fraction of agents are classified as \textit{Routiners} despite remaining at home for nearly the entire simulation, exhibiting maximum \textit{stationarity} and near-zero \textit{regularity}, i.e.,
a degenerate routine pattern rather than realistic routine behavior.

\subsection{Semantic Mobility Dynamics}

\begin{figure}[!htb]
    \centering
    \begin{subfigure}{0.43\columnwidth}
        \centering
\includegraphics[width=\linewidth]{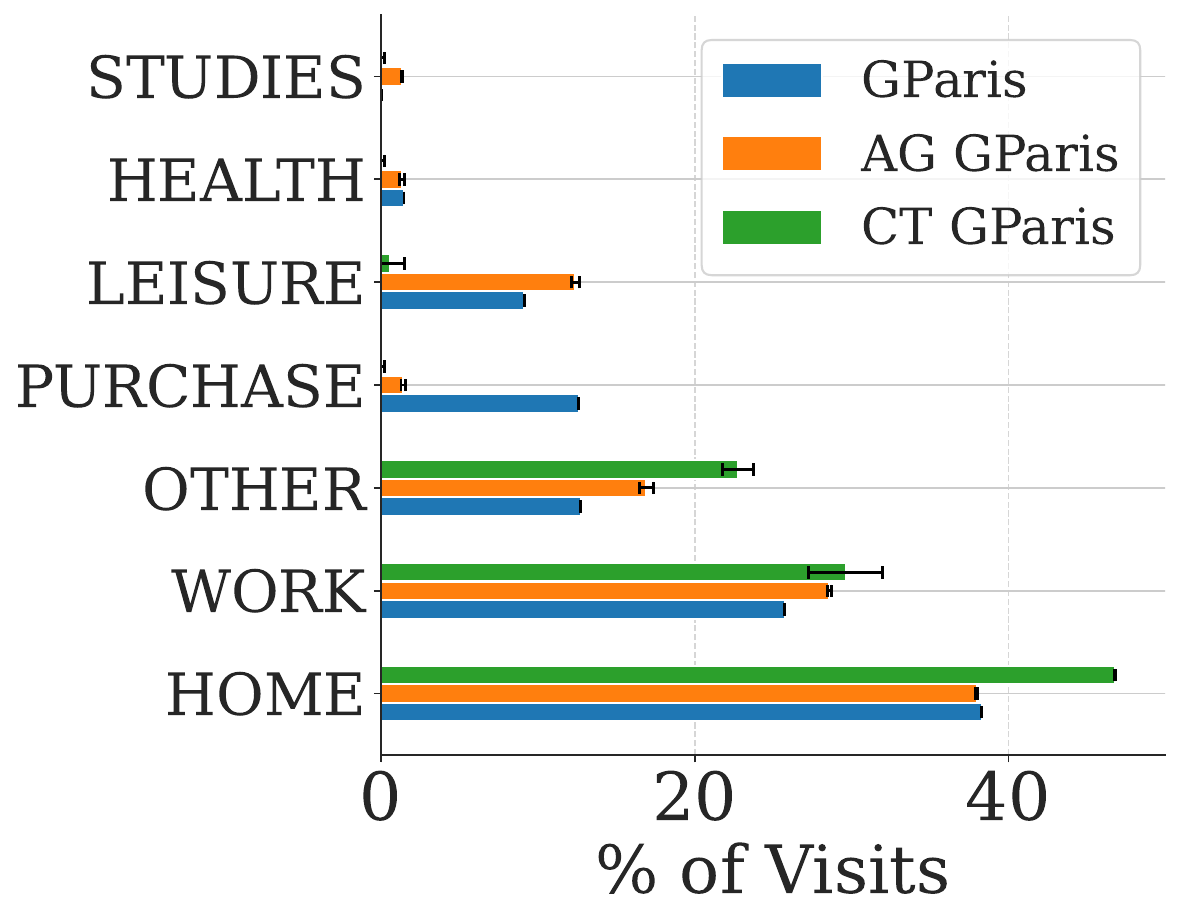}
        \caption{\textit{Visit Purpose Distribution}}
        \label{fig:semantic_vpd}
    \end{subfigure}
    \hfill
    \begin{subfigure}{0.5\columnwidth}
        \centering
\includegraphics[width=\linewidth]{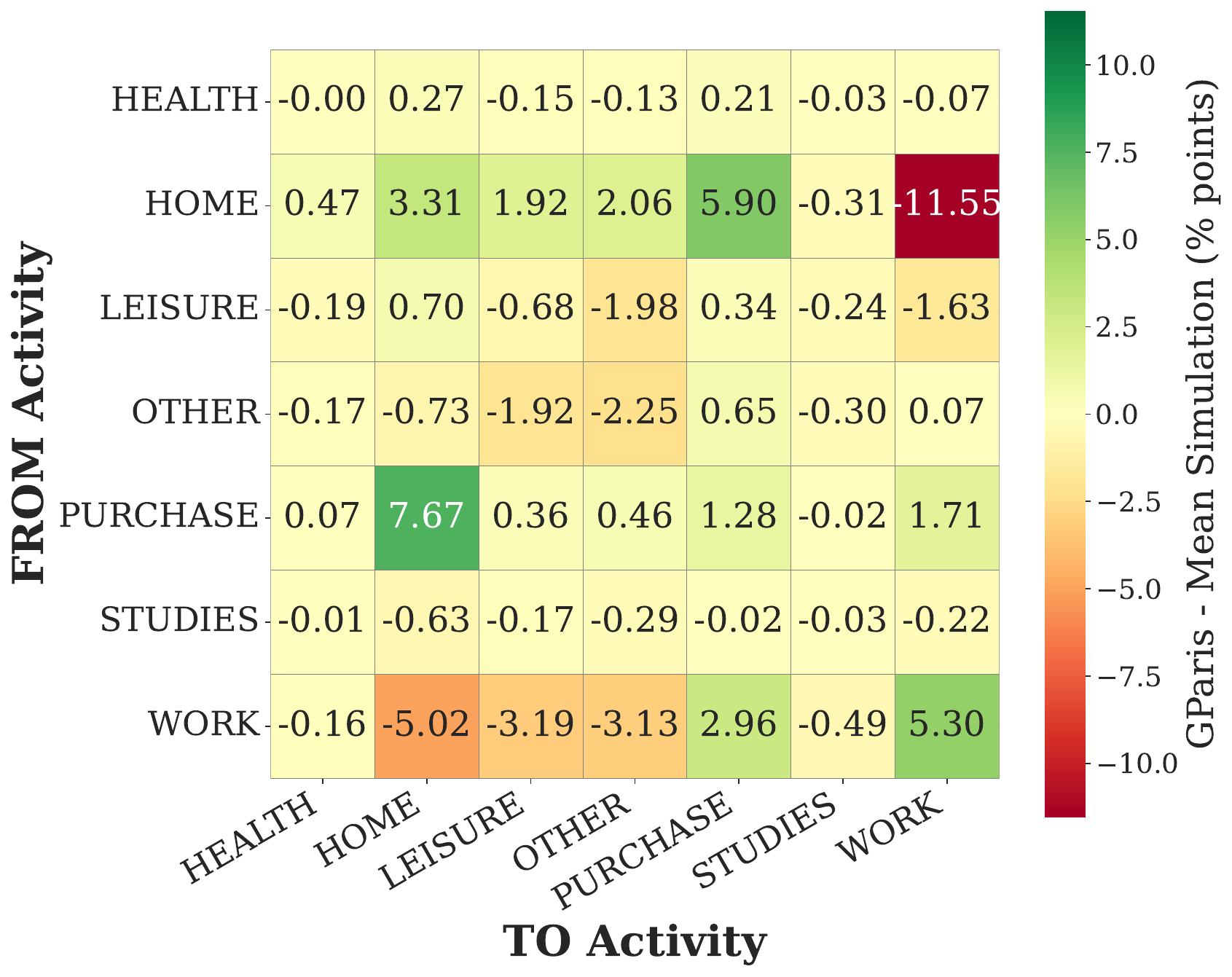}
    \caption{\textit{Activity Transition Matrix}}
        \label{fig:semantic_atm}
    \end{subfigure}\vspace{-0.3cm}
    \caption{Semantic mobility patterns for the \parisDataShort{}. 
    } 
    \label{fig:semantic}
\end{figure}

\greybox{\textit{LLM-based agents reproduce first-order semantic activity distributions more reliably than temporal and sequential semantic dynamics.} } \vspace{-0.3cm}
Fig.~\ref{fig:semantic} summarizes semantic activity behavior in the \parisData{} through aggregate visit-purpose distributions and differences in activity-transition probabilities, while Table~\ref{tab:semantic_results} reports the corresponding \textit{JSD} metrics. We report semantic metrics only for the \parisData{}, since the Shanghai dataset does not include activity-purpose annotations.

\begin{table}[t]
\centering
\scriptsize
\caption{Semantic mobility realism metrics comparing empirical and simulated trajectories. } 
\vspace{-0.3cm}
\label{tab:semantic_results}
\begin{tabular}{llccc}
\toprule
\textbf{Dataset} & \textbf{Source} & \textbf{\textit{VPD}} (\textit{JSD}) & \textbf{\textit{ATM}} (\textit{JSD}) & \textbf{\textit{DARD}} (\textit{JSD}) \\
\midrule

\parisDataShort{} & \textit{AgentSociety} & $0.0296 \pm 0.0010$ & $0.1125 \pm 0.0029$ & $0.0110 \pm 0.0004$ \\

\parisDataShort{} & \textit{CitySim} & $0.0883 \pm 0.0255$ & $0.2405 \pm 0.0559$ & $0.1573 \pm 0.2156$ \\

\bottomrule
\end{tabular}
\end{table}

For the \parisData{}, \textit{AgentSociety} produces semantic activity proportions broadly matching the empirical data, despite generating fewer visits overall. This is reflected in its low \textit{VPD} divergence ($0.0296 \pm 0.0010$). The main discrepancy occurs in purchase-related activities, which are underrepresented in the simulation. This mismatch may reflect limitations in POI classification, POI availability, or destination-selection behavior within the simulation.

Compared to \textit{CitySim}, \textit{AgentSociety} more closely matches empirical semantic behavior across all three metrics. Although
\textit{CitySim} introduces more complex destination-selection mechanisms, these additional constraints may reduce robustness when compatible POIs are sparse or unavailable, reducing semantic activity accuracy.

\begin{figure}[!tbh]
    \centering
    \includegraphics[width=.8\linewidth]{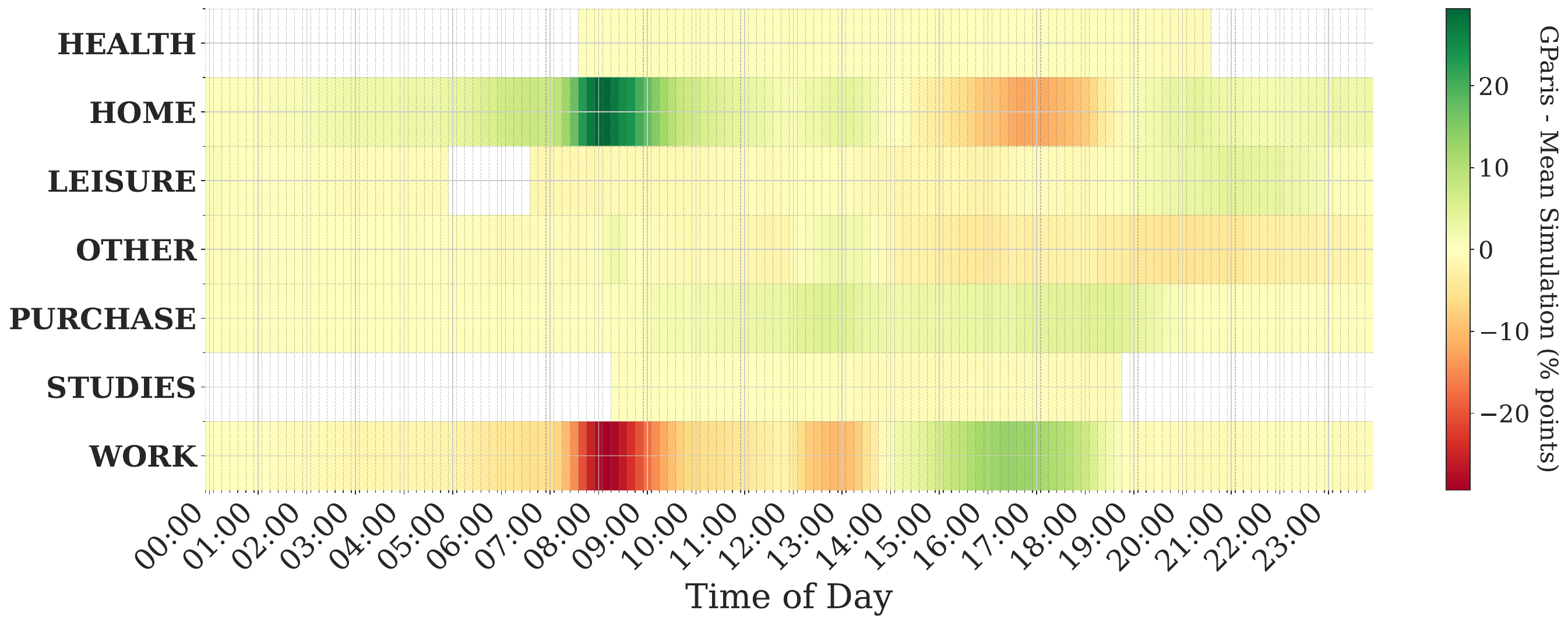}\vspace{-0.3cm}
    \caption{\textit{Daily Activity Routine Distribution} (\textit{DARD}) differences for the \parisData{}.
    }
    \label{fig:DARD}
\end{figure}

Fig.~\ref{fig:DARD} further shows that simulated agents partially reproduce coarse daily activity rhythms throughout the day, although some activity purposes remain over- or under-represented at specific time intervals. However, the larger \textit{ATM} divergence indicates that realistic aggregate activity proportions do not necessarily translate into realistic activity sequences.

\greybox{\textit{Generating plausible individual activities is easier than reproducing the ordered transitions through which these activities occur in empirical mobility.}}

\section{Toward High-Fidelity Urban Simulation}
\label{sec:discussion}

This section discusses potential directions for improving the mobility realism limitations identified in LLM-based urban simulators.

\subsection{Improving Spatial and Semantic Realism}

\noindent\textit{\textbf{Improving POI Representation.}} Our analysis suggests that part of the observed spatial and semantic mismatch originates from limitations in the underlying POI representation. In Greater Paris area, 57.45\% of the 223,149 OpenStreetMap (OSM) POIs correspond to benches, bicycle parking, waste baskets, recycling points, or post boxes, leaving fewer than 100,000 semantically meaningful destinations for agent decision-making.

To evaluate if richer POI coverage improves simulation realism, we augmented the map with 178,884 deduplicated Overture Maps POIs~\cite{overture_maps_2026}. We then compared semantic POI density at H3 resolution 8 against the official French facility inventory, \textit{Base Permanente des Équipements} (BPE)~\cite{inseeBPE}. The Pearson correlation increased from $0.55 \pm 0.23$ using OSM data alone to $0.80 \pm 0.20$ using OSM+Overture. 
\greybox{\textit{Results suggest that improving semantic map coverage may substantially improve destination selection and activity realism in LLM-based urban simulators.}}\vspace{-0.2cm}

\noindent\textit{\textbf{Transport Mode and Routing Realism.}} Transport behavior also plays a central role in mobility realism. In the original \textit{AgentSociety} mobility layer, car-based routing relies on simplified travel-time assumptions, including a fixed-duration fallback, rather than context-sensitive travel times. This simplification directly affects \textit{$\Delta r$} distributions, \textit{dwell time}, \textit{visitation frequency}, and temporal activity rhythms.

Although \textit{CitySim} introduces transport-mode selection mechanisms \cite{citysim25}, reproducing its full routing behavior remains difficult because \textit{AgentSociety} depends on a partially closed-source binary traffic layer \cite{agentsociety_mobilitysystem23}. We provide an open reimplementation of this traffic layer for car-based trips for comparability; routing for other transport modes remains future work.
\greybox{\textit{Future urban simulators should integrate fully open and multimodal routing engines capable of modeling realistic travel times across walking, rail, cycling, and vehicular transportation systems.}}

\subsection{Advancing Behavioral and Social Dynamics}

\noindent\textit{\textbf{Explicit Mobility Profiling.}} Our results suggest that realistic behavioral \textit{diversity} does not reliably emerge solely from generic prompting. Incorporating behavioral constraints may improve individual mobility realism and aggregate population structure.
\greybox{\textit{Future simulation frameworks should 
incorporate explicit mobility-behavior mechanisms capable of modeling heterogeneous exploration and return dynamics across individuals. Rather than assuming patterns naturally emerge from high-level personas or preferences, simulators could directly condition agent behavior on mobility characteristics such as exploration persistence, routine stability, preferred \textit{motif} complexity, or the tendency to revisit familiar locations.}} \vspace{-0.2cm}  

\noindent\textit{\textbf{Social Network Effects.}} Social interactions also influence mobility behavior through group activities, co-location patterns, and new places' discovery. However, the initialization and evolution of social relationships remain poorly documented in current LLM-based urban simulators such as \textit{AgentSociety}~\cite{agentsociety} and \textit{CitySim}~\cite{citysim25}. Because our evaluation focused primarily on mobility realism, we did not systematically analyze the impact of social-network structure. 
\greybox{\textit{Future work should investigate how different social-network initialization strategies influence exploration behavior, routine formation, and collective mobility dynamics.}}

\subsection{Computational Efficiency and Open Science}

\textit{\textbf{Computational Cost and Scalability.}} Large-scale LLM-based urban simulation remains computationally expensive. Consequently, several mobility laws and population-level dynamics could not be evaluated at their full scale, as our experiments were restricted to populations of approximately 500 agents. Table~\ref{tab:agentsociety_efficiency_cost} in Appendix~\ref{sec:appdx_costs}, shows that simulating 7--10 days of activity for 500 agents with GPT-4o-mini, i.e., the same model used in CitySim~\cite{citysim25}, would cost roughly \$130–200. These simulations consume 700–900 million input tokens and 65–110 million output tokens, resulting in execution times of several days even at this limited scale.

\greybox{\textit{Future systems will likely require hierarchical or hybrid architectures that combine LLM reasoning with lightweight specialized models. Routine tasks such as dispatcher classification, regression updates in NeedsBlock, or repetitive behavioral decisions could perhaps be handled by fine-tuned compact models or other more efficient alternatives, reserving larger LLM calls for planning, reasoning, and complex social interactions.}} \vspace{-0.2cm}

\noindent\textit{\textbf{Observability and Reproducibility.}} Transparent and reproducible infrastructure is essential for evaluating LLM-based urban simulators. Closed-source systems, as \textit{CitySim}, limit external validation and debugging, while open platforms, as \textit{AgentSociety}, enable inspection of cognitive and mobility-generation mechanisms. \textit{Nevertheless, effective observability requires more than source-code access alone;} the original \textit{AgentSociety} implementation provided only limited visibility into the internal state and decisions driving agent behavior, motivating enhancements introduced in this work. 

\greybox{\textit{For reproducibility and debugging, we release an anonymized fork of \textit{AgentSociety} with built-in observability and support for switching between \textit{AgentSociety} and \textit{CitySim} configurations. }}\vspace{-0.2cm} 

The framework\footnote{\ArtifactURL} records visited POIs, semantic categories, prompts and responses, memories, emotions, attitudes, thoughts, block execution times, LLM latency, token consumption, and call frequencies. 
\greybox{\textit{Beyond enabling detailed cost and behavior analysis, this state tracking also improves fault tolerance by supporting checkpoint-based recovery, allowing long-running simulations to resume from the last saved state after failures, rather than restarting from the beginning. 
}}\vspace{-0.2cm}

\noindent\textit{\textbf{Open Mobility Evaluation Infrastructure.}} We additionally release four complementary anonymized artifacts. First, a scalable regional map generation tool. Second, we release \textit{En-AgentSociety}\footnote{\TrafficSimArtifactURL}, an extended and fully reproducible version of \textit{AgentSociety} that integrates our reimplementation of both \textit{AgentSociety} and \textit{CitySim} within a unified simulation artifact. \textit{En-AgentSociety} also reimplements modules supporting future studies of multimodal mobility and social interactions (see Appendix \ref{appendixMulti}). Third, a Rust+Python evaluation framework\footnote{\MetricsArtifactURL} computes all mobility and semantic metrics used in this work, including spatial, temporal, network, behavioral-profile, and POI-transition measures. Fourth, an open reimplementation of the traffic simulator \textit{AgentSociety} and \textit{CitySim} uses. This framework facilitates debugging of routing and mobility behavior while establishing a public foundation for future multimodal transportation support.

\section{Conclusion}
\label{sec:conclusion}

This work introduced an empirical mobility realism evaluation framework for LLM-driven urban simulators grounded in real-world mobility data. Using datasets from Greater Paris area and Shanghai, we evaluated \textit{AgentSociety} and \textit{CitySim} across multiple dimensions of mobility realism, including spatial mobility laws, temporal dynamics, topological motifs, semantic activity behavior, and behavioral mobility profiles.

Our results show that current LLM-based simulators can reproduce portions of routine human behavior, particularly coarse semantic activity distributions and simple repetitive mobility structures. However, substantial discrepancies remain in large-scale spatial flows, temporal rhythms, sequential activity transitions, and heterogeneous exploration behavior. These findings suggest that generating plausible individual trajectories is fundamentally different from reproducing empirically grounded urban mobility dynamics, highlighting the need for more observable, scalable, and behaviorally grounded urban simulation systems.

This study has two main limitations. First, because \textit{CitySim} is not publicly available, our evaluation relies on a reimplementation based on the functionalities described in the original paper; implementation differences may therefore affect the observed results. Second, semantic mobility evaluation is restricted to the \parisData{}, since the Shanghai dataset does not include activity-purpose annotations. Future work should evaluate additional public simulators, richer semantic mobility datasets, and stronger spatial constraints for LLM-based mobility generation.

\begin{acks}
This study was supported by CNPq (processes 314603/2023-9, 441444/2023-7, and 444724/2024-9) - and the PEPR MOBIDEC Mob Sci-Dat Factory project. This research is also part of the INCT TILD-IAR funded by CNPq (proc. 408490/2024-1). 
\end{acks}

\clearpage
\bibliographystyle{ACM-Reference-Format}

\clearpage 

\appendix

\begin{table*}[th]
\centering
\scriptsize
\caption{Mobility metrics and laws used to validate LLM-based urban simulators against empirical human mobility behavior.}\vspace{-0.3cm} 
\label{tab:methodology_metrics}
\begin{tabular}{p{0.08\textwidth}p{0.13\textwidth}p{0.23\textwidth}p{0.45\textwidth}}
\toprule
\textbf{Type} & \textbf{Laws / Metric} & \textbf{Formulation} & \textbf{Intuition / Interpretation} \\
\midrule
\multirow{6}{0.12\textwidth}{Spatial laws and empiriral scaling \\ laws}
& Travel distance ($\Delta r$)~\cite{Gonzalez2008} & $\Delta r_i = ||x_{i+1} - x_i||$; $P(\Delta r)=(\Delta r+\Delta r_0)^{-\beta}e^{-\Delta r/\kappa}$ & Captures distances between consecutive locations. Real mobility follows a truncated power-law distribution, so deviations indicate unrealistic displacement scales. \\
& Radius of Gyration ($r_g$)~\cite{Gonzalez2008} & $r_g=\sqrt{\frac{1}{n}\sum_i\left\lVert l_i-l_{\mathrm{cm}}\right\rVert^2}$ & Measures each individual's characteristic spatial range around their center of mass, capturing whether simulated agents remain realistically spatially bounded. \\
& Daily visits~\cite{Schneider2013} & $N_{day}=\left|\{location_i:t_i\in day\}\right|$ & Captures the number of distinct locations visited per day, which empirically follows a log-normal-like distribution. \\
& Predictability~\cite{Song2010Entropy} & $\Pi^{\max}$ from $S=-\Pi\log_2\Pi-(1-\Pi)\log_2(1-\Pi)+(1-\Pi)\log_2(N-1)$ & Estimates the upper bound of mobility predictability from trajectory entropy, indicating whether simulated mobility is too regular or too random. \\
& Distance-frequency ~\cite{Schlapfer2021-pw} & $\rho_i(r,f)=\frac{\mu_i}{(rf)^\eta}$ & Captures the joint relationship between travel distance from home and visitation frequency, distinguishing frequent short-range visits from infrequent long-range trips. \\
& OD matrix \cite{Barbosa2018} & $F_{ab} = \sum_i \mathbf{1}_{(a \to b)_i}$ & Measures agreement between empirical and simulated origin-destination movement flows at multiple H3 resolutions. \\
\midrule
\multirow{3}{0.12\textwidth}{Temporal  metrics}
& Trip duration (TD.) \cite{Barbosa2018}  & $\tau_i=t^{\mathrm{arrival}}_{i+1}-t^{\mathrm{departure}}_i$ & Captures the distribution of travel times between consecutive visits, reflecting routing, transportation, and distance-generation realism. \\
& Dwell time (DT) \cite{Barbosa2018}  & $\delta_i=t^{\mathrm{departure}}_i-t^{\mathrm{arrival}}_i$ & Measures how long agents remain at locations before moving again, capturing activity-duration realism. \\
& Visitation frequency (Vf.) \cite{Schneider2013} & $f_l=\sum_i\mathbf{1}[l_i=l]$ & Captures how often users visit locations over the analysis period, exposing under- or over-generation of mobility events. \\
\midrule
\multirow{1}{0.12\textwidth}{Topological laws}
& Mobility motifs~\cite{Schneider2013} & $G_d=(V_d,E_d)$ from daily location sequence; motif $=$ graph isomorphism class & Represents daily routines as directed graphs with up to six nodes, capturing recurrent topological structures in everyday mobility. \\
\midrule
\multirow{5}{0.12\textwidth}{Behavioral metrics}
& Mobility profiles~\cite{licia_sig2020,esper_sig24} & $\mathrm{profile}=\mathrm{GMM}(\mathrm{Intermittency},\mathrm{Deg.\ of\ Return})$ & Classifies individuals as \textit{Scouters}, \textit{Regulars}, or \textit{Routiners} from exploration and return dynamics. \\
& Regularity~\cite{Teixeira2021,teixeira:hal-03128639} & $\mathrm{Reg.}=\frac{1}{T}\sum_t\mathbf{1}[l_t\in L_{\mathrm{known}}]$ & Measures the tendency to repeatedly revisit the same locations. \\
& Stationarity~\cite{Teixeira2021,teixeira:hal-03128639} & $\mathrm{Stat.}=\frac{1}{T-1}\sum_t\mathbf{1}[l_t=l_{t+1}]$ & Captures the frequency with which users remain in the same location across consecutive observations. \\
& Diversity~\cite{Teixeira2021} & $\mathrm{Div.}=\left|\{\mathrm{subtrajectory\ patterns}\}\right|$ & Quantifies the variety of observed sub-trajectories or location-sequence patterns in an individual's mobility behavior. \\
& Entropy~\cite{Song2010Entropy,Teixeira2019,teixeira:hal-03128639} & $S_{\mathrm{real}}\approx\frac{n\log_2(n)}{\sum_{i\le n}\Lambda_i}$ & Measures the uncertainty and temporal complexity of trajectories, complementing profile-level exploration behavior. \\
\midrule
\multirow{4}{0.12\textwidth}{Semantic and spatio-temporal \\laws}
& Visit Purpose Distribution (VPD) \cite{Widhalm2015} & $P(c)=\frac{1}{N}\sum_i\mathbf{1}[c_i=c]$ & Captures the overall distribution of activity categories, such as home, work, shopping, and leisure. \\
& Activity Transition Matrix (ATM)~\cite{senefonteSocInfo2020} & $P(c_j\mid c_i)=\frac{\#(c_i\rightarrow c_j)}{\sum_k\#(c_i\rightarrow c_k)}$ & Measures sequential transitions between activity types, testing whether plausible activities occur in realistic orders. \\
& Daily Activity Routine Distribution (DARD)~\cite{wang2024largelanguagemodelsurban} & $P(c,t)=\frac{1}{N}\sum_i\mathbf{1}[c_i=c,t_i=t]$ & Captures activity distributions across 10-minute intervals throughout the day, measuring temporal semantic rhythms. \\
& Spatio-Temporal Visit Distribution (STVD)~\cite{wang2024largelanguagemodelsurban} & $P(c,h,t)=\frac{1}{N}\sum_i\mathbf{1}[c_i=c,h_i=h,t_i=t]$ & Extends DARD by jointly modeling visits across time intervals and spatial regions, such as H3 cells. \\
\midrule
\multirow{3}{0.12\textwidth}{Similarity metrics}
& Jensen-Shannon Divergence (JSD) \cite{jsd_paper} & $\mathrm{JSD}(P,Q)=\frac{1}{2}D_{\mathrm{KL}}(P\Vert M)+\frac{1}{2}D_{\mathrm{KL}}(Q\Vert M)$, $M=\frac{P+Q}{2}$ & Compares categorical distributions, including VPD, ATM, DARD, STVD, motifs, and profile distributions. \\
& Common Part of Commuters (CPC) \cite{Barbosa2018} & $\mathrm{CPC}=\frac{2\sum_{ab}\min(F_{ab},\hat{F}_{ab})}{\sum_{ab}F_{ab}+\sum_{ab}\hat{F}_{ab}}$ & Evaluates overlap between empirical and simulated OD matrices; higher values indicate better flow agreement. \\
& Wasserstein distance ($W_1$) \cite{villani_topics_2003} & $W_1(P,Q)=\inf_{\gamma\in\Gamma(P,Q)}\mathbb{E}_{(x,y)\sim\gamma}[d(x,y)]$ & Compares ordinal or numerical distributions, including travel distance, trip duration, \textit{radius of gyration}, dwell time, and visitation frequency. \\
\bottomrule
\end{tabular}
\end{table*}

\section{Map Generation Scalability Benchmark}
\label{sec:appdx_mosstool_benchmark}

We benchmark our optimized map-generation pipeline against the original MOSSTool. As discussed in \textbf{\S\ref{sec:sim_configuration}}, MOSSTool struggles with large regions due to costly spatial matching and memory-intensive serialization. Our spatial indexing and batch-processing optimizations reduce runtime and memory use, enabling regional-scale map generation. Table~\ref{tab:mosstool_benchmark} reports results for Massy, a representative urban subset in Greater Paris area.

\begin{table}[htb]
\centering
\caption{Map generation performance for Massy (GP).}
\vspace{-0.3cm}
\label{tab:mosstool_benchmark}
\footnotesize
\begin{tabular}{lcc}
\toprule
\textbf{Implementation} & \textbf{Execution Time (s)} & \textbf{Peak Memory (MB)} \\
\midrule
Original MOSSTool & 135.86 $\pm$ 5.99 & 633.63 $\pm$ 0.21  \\
Optimized Pipeline (Ours) & 63.87 $\pm$ 0.54 & 546.92 $\pm$ 50.23 \\
\bottomrule
\end{tabular}
\end{table}

\section{\textit{CitySim} Reimplementation Choices}
\label{sec:appdx_citysim}

Next, we summarize the choices used to reimplement the 
\textit{CitySim} reported functionalities within our \textit{AgentSociety}-based artifact.

\subsection{Persona Module}

We extend the \textit{AgentSociety} persona with the attributes described in \textit{CitySim}: life stage, Big Five traits, habits/hobbies, and preferences. Since \textit{CitySim} derives these attributes from questionnaires, we approximate them from the agent attributes already available in \textit{AgentSociety}'s synthetic personas using LLM prompting.

\subsection{Memory Module}

We model \textit{CitySim}'s spatial memory as POI-level beliefs over price, atmosphere, satisfaction, and convenience. Unvisited POIs are initialized from embedding similarity to previously visited locations with uncertainty $u_0 = 0.25$.

After each visit, the \texttt{NeedsBlock} updates the agent's needs, and an LLM estimates the four subjective beliefs from the agent profile and POI context. Spatial memory is updated with a 1D Kalman filter using fixed observation noise $\sigma_{\text{obs}} = 0.2$, with each belief updated as $K \cdot \text{observation} + (1-K)\cdot\text{prior}$, where $K$ depends on current uncertainty. Updated POIs are re-embedded, and beliefs decay daily toward a neutral baseline with $\alpha_{\text{decay}} = 0.03$.

\subsection{Long-Term Goal Module}

The long-term goal module generates and revises agent aspirations from persona, financial status, recent activity, social context, and previously generated goals. We compute the prompt-level behavioral indicators described in \textit{CitySim}: need fulfillment, financial stress, social isolation, and interest in recently visited POIs. Need fulfillment is the share of the day in which core needs remain above target thresholds. As in \textit{CitySim}, goals are updated monthly.

\subsection{Planning Module}

We reimplement \textit{CitySim}'s planning mechanics through a custom \texttt{DailyScheduleBlock} integrated into \textit{AgentSociety}'s \texttt{Forward} pass, preserving the simulator's memory and environment interfaces.

Planning has two phases. At the start of each simulated day, the block queries the LLM for a structured JSON schedule using the agent profile, Big Five traits, preferences, chronotype, and current need satisfaction. Following \textit{CitySim}'s recursive decomposition strategy, high-priority tasks such as sleep and work are scheduled before medium-priority tasks such as meals and hygiene; remaining intervals are marked as \texttt{[EMPTY]} for leisure or long-term goals.

When simulation time reaches an \texttt{[EMPTY]} block, the module triggers value-driven execution: it retrieves the agent's location, emotional state, recent thoughts, and urgent needs, then asks the LLM to generate and evaluate 2--4 candidate activities. The selected activity maximizes expected satisfaction gain under the Maslow-style need hierarchy.

\subsection{Destination Selection Module}

We implement belief-aware destination selection through a multi-stage \texttt{PlaceSelectionBlock}, which adds a Neighborhood-level spatial hierarchy to \textit{AgentSociety}'s map queries and falls back to AOIs when neighborhood data are unavailable.

\textbf{Macro-level area selection:} When an agent forms an intention, the LLM predicts primary/secondary POI categories and an acceptable travel radius from risk tolerance, emotional state, and weather. Matching POIs are mapped to enclosing Neighborhood polygons, and the LLM selects 3--5 candidate Neighborhoods using the persona, 7-day visit history, and matching-POI density. If this fails, the block falls back to AOI-level selection ranked by distance and popularity.

\textbf{Micro-level POI selection:} Within selected Neighborhoods or AOIs, candidate POIs are filtered and ranked using a belief-weighted gravity model. We average the four spatial-memory beliefs into score $b_j$ and use distance decay $distance^{1+\gamma(b_j-0.5)}$, with $\gamma=2.0$, allowing highly valued POIs to attract longer trips while preserving distance sensitivity. The selected POI is passed to \texttt{MoveBlock}.

\subsection{Multimodal and Social Modules}\label{appendixMulti}

Although not evaluated in this work, we reimplement Multimodal and Social modules in \textit{En-AgentSociety} to support future studies of multimodal mobility and social interactions. We exclude them from our experiments because reproducible multimodal routing and realistic social-network initialization remain insufficiently supported in the available simulator infrastructure.

\section{Mobility Metrics and Laws}\label{sec_appendix_metrics}

Table \ref{tab:methodology_metrics} presents all metrics and laws used in the study.

\section{Spatial Mobility-Law Population Comparisons}
\label{sec:appx_spatial_cdr_population}

\begin{figure}[tb]
    \centering
    \includegraphics[width=1\linewidth]{figures/line_legend.png}
    \centering
    \subfloat[\textit{$\Delta r$}]{
        \includegraphics[width=0.15\textwidth]{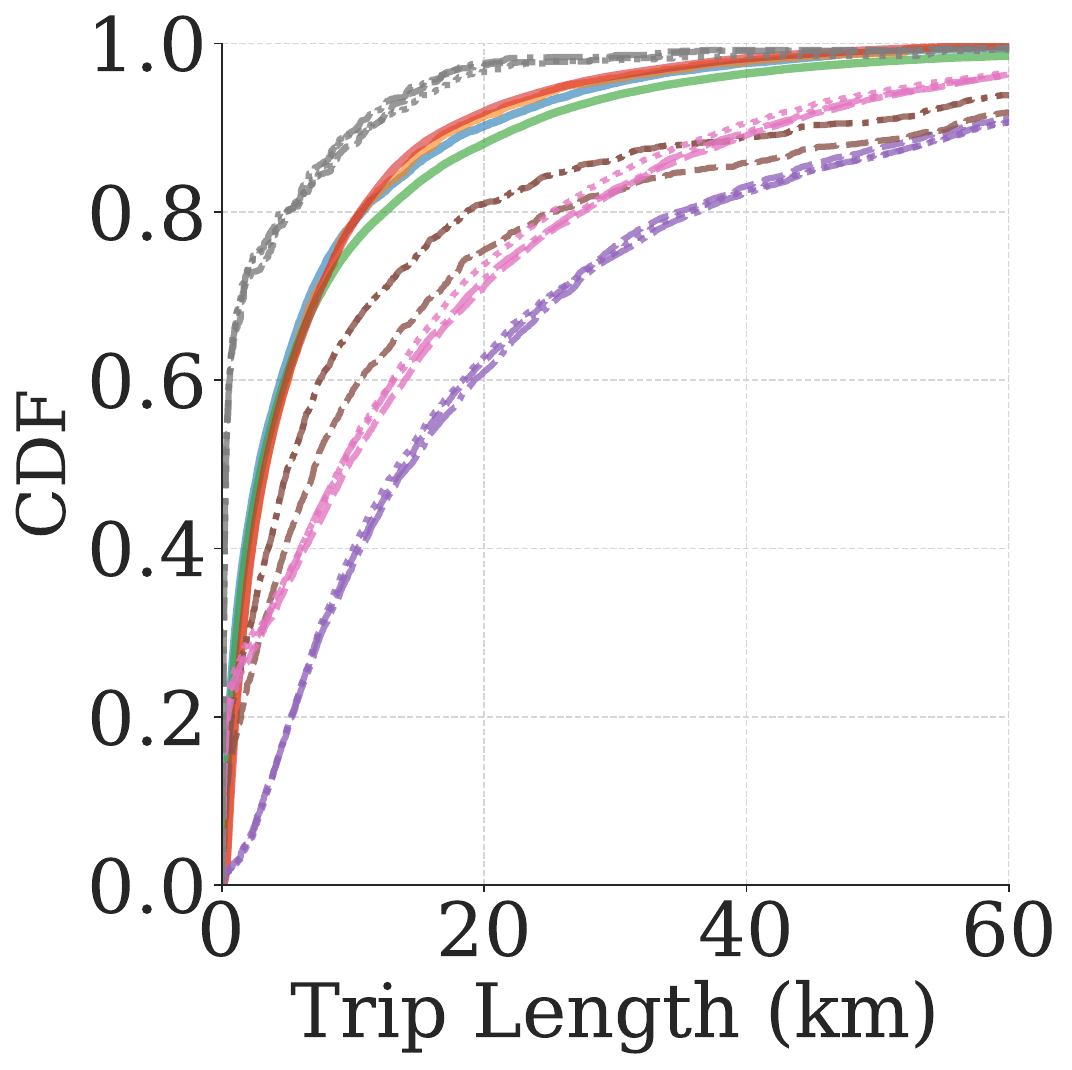}
    }
    \hfill
    \subfloat[\textit{$r_g$}]{
        \includegraphics[width=0.15\textwidth]{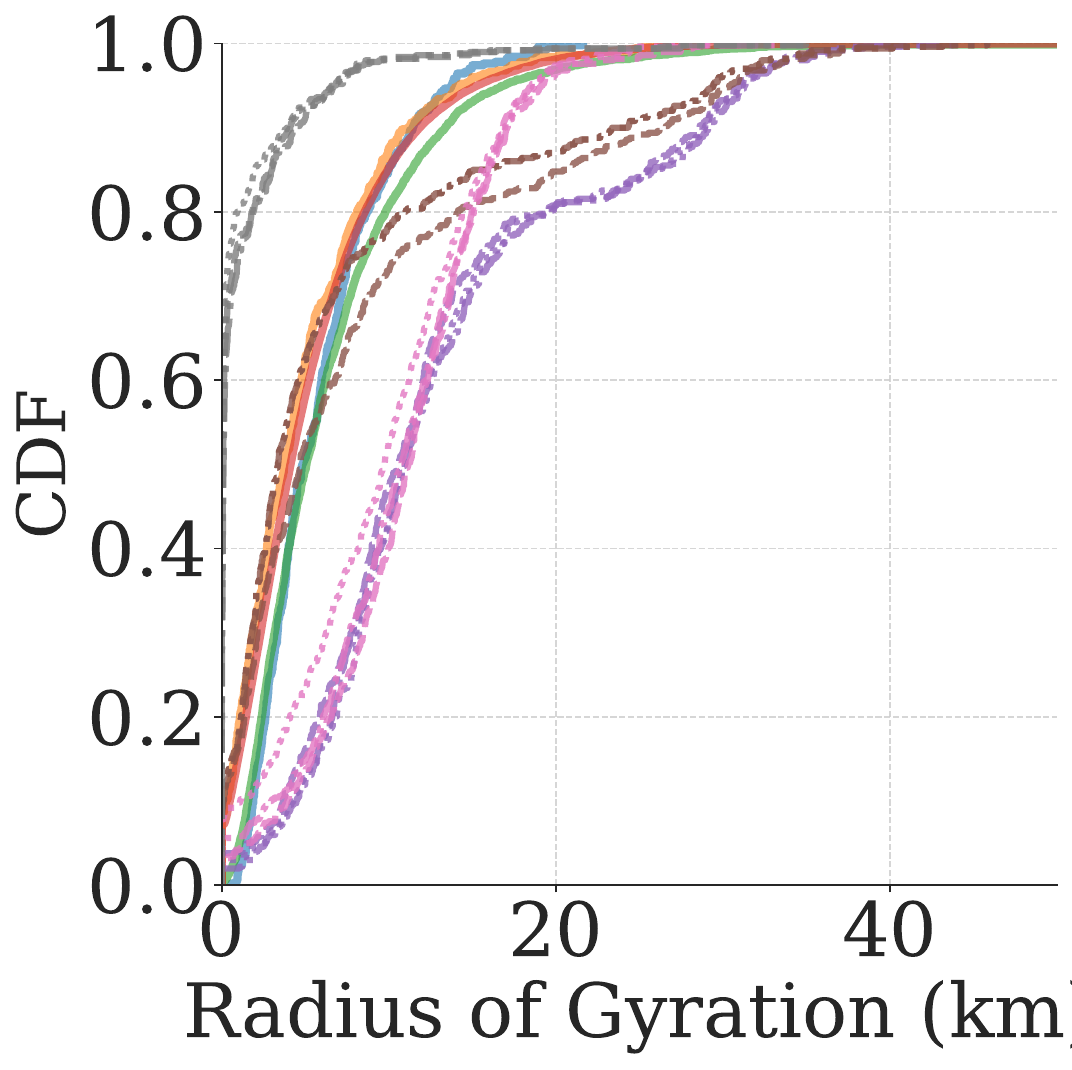}
    }\vspace{-0.3cm}
    \caption{CDFs of \textit{travel distance} ($\Delta r$) and \textit{radius of gyration} ($r_g$) for empirical samples, full populations,
and simulated trajectories.}
    \label{fig:spatial_population_cdfs}
\end{figure}

We compare spatial mobility-law distributions across simulated trajectories, the 500-user empirical samples, and the full empirical populations. Since overlaying all groups in the mobility-law plots reduces readability, we report CDFs for \textit{travel distance} ($\Delta r$) and \textit{radius of gyration} ($r_g$) -- see Figure \ref{fig:spatial_population_cdfs}.

\section{Computational Costs}
\label{sec:appdx_costs}

\begin{table}[tb] 
\centering 
\caption{Estimated API costs for simulation runs. Costs use GPT-4o-mini pricing following the \textit{CitySim} configuration~\cite{citysim25}: $0.15\$/M$ input tokens and $0.60\$/M$ output tokens. Values provide a normalized cost comparison rather than actual billing for all configurations.}
\vspace{-0.3cm}
\label{tab:agentsociety_efficiency_cost} 
\footnotesize 
\begin{tabular}{lcccc} 
\toprule 
\textbf{Dataset / Source} & \textbf{\# Agents} & \textbf{\# Days} & \textbf{In/Out Tokens} & \textbf{Cost (USD)} \\ 
\midrule 
\parisDataShort{} (AG) & 504 & 7  & 720.7 M / 110.9 M & \$174.65 \\ 
Shanghai (AG)      & 500 & 10 & 887.8 M / 97.5 M  & \$191.67 \\ 
\parisDataShort{} (CT) & 504 & 7  & 609.6 M / 66.7 M  & \$131.46 \\ 
Shanghai (CT)      & 500 & 10 & 950.9 M / 106.6 M & \$206.60 \\ 
\bottomrule 
\end{tabular} 
\end{table}

Table \ref{tab:agentsociety_efficiency_cost} reports token usage and estimated API costs for the simulation runs.

\end{document}